\documentclass[journal]{IEEEtran}
\usepackage{ifpdf}
\usepackage{cite}
\usepackage{xcolor}
\ifCLASSINFOpdf
   \usepackage[pdftex]{graphicx}
\else
   \usepackage[dvips]{graphicx}
   \graphicspath{{../eps/}}
\fi

\usepackage{amsmath}
\usepackage{amssymb}
\usepackage{amsfonts}
\usepackage{algorithmic}
\usepackage{multirow}
\usepackage{array}
\usepackage{hhline}
\usepackage{mathrsfs}
\usepackage{booktabs}
\ifCLASSOPTIONcompsoc
 \usepackage[caption=false,font=normalsize,labelfont=sf,textfont=sf]{subfig}
\else
 \usepackage[caption=false,font=footnotesize]{subfig}
\fi

\usepackage{fixltx2e}

\usepackage{stfloats}

\usepackage{url}

\def\eg{e.g.,~}
\def \NOTE [#1]{\textcolor{blue}{(\textit{#1})}}

\begin{document}

\title{Conceptual Compression via Deep Structure and Texture Synthesis}

\def\etal{\textit{et~al}.\xspace}
\def\ie{\textit{i.e.},\xspace}

\author{Jianhui~Chang, Zhenghui~Zhao, Chuanmin~Jia, Shiqi~Wang, Lingbo~Yang, Qi Mao, Jian~Zhang and Siwei~Ma

\thanks{Manuscript received August 27, 2020; revised October 28, 2021; accepted March 1, 2022. This work was supported in part by the National Natural Science Foundation of China (62025101, 62101007, and 62088102); in part by the National Postdoctoral Program for Innovative Talents under Grant BX2021009; and in part by the High Performance Computing Platform of PKU. (\textit{Corresponding author: Siwei Ma, Jian Zhang.})}
\thanks{Jianhui Chang, Chuanmin Jia, Lingbo Yang and Siwei Ma are with National Engineering Research Center of Visual Technology, School of Computer Science, Peking University, Beijing 100871, China. Siwei Ma is also with the Information Technology Research and Development Innovation Center, Peking University, Shaoxing 312000. (e-mail: jhchang@pku.edu.cn; cmjia@pku.edu.cn; lingbo@pku.edu.cn; swma@pku.edu.cn).}
\thanks{Jian Zhang is with the School of Electronic and Computer Engineering, Peking University Shenzhen Graduate School, Shenzhen 518055, China and is also with the Peng Cheng Laboratory, Shenzhen,
China. (e-mail: zhangjian.sz@pku.edu.cn).}
\thanks{Shiqi Wang is with the Department of Computer Science, City University of Hong Kong, Hong Kong. (e-mail: shiqwang@cityu.edu.hk).}
\thanks{Zhenghui Zhao is with LMAM, School of Mathematical Sciences, Peking University, Beijing 100871, China. (e-mail: zhzhao@pku.edu.cn).}
\thanks{Qi Mao is with School of Information and Communication Engineering, the State Key Laboratory of Media Convergence and Communication, Communication University of China, Beijing 100024, China. (e-mail: qimao@cuc.edu.cn).}
}



\maketitle

\begin{abstract}

Existing compression methods typically focus on the removal of signal-level redundancies, while the potential and versatility of decomposing visual data into compact conceptual components still lack further study.
To this end, we propose a novel conceptual compression framework that encodes visual data into compact structure and texture representations, then decodes in a deep synthesis fashion, aiming to achieve better visual reconstruction quality, flexible content manipulation, and potential support for various vision tasks.
In particular, we propose to compress images by a dual-layered model consisting of two complementary visual features: 1) structure layer represented by structural maps and 2) texture layer characterized by low-dimensional deep representations.
At the encoder side, the structural maps and texture representations are individually extracted and compressed, generating the compact, interpretable, inter-operable bitstreams.
During the decoding stage, a hierarchical fusion GAN (HF-GAN) is proposed to learn the synthesis paradigm where the textures are rendered into the decoded structural maps, leading to high-quality reconstruction with remarkable visual realism. 
Extensive experiments on diverse images have demonstrated the superiority of our framework with lower bitrates, higher reconstruction quality, and increased versatility towards visual analysis and content manipulation tasks.

\end{abstract}


\IEEEpeerreviewmaketitle

\section{Introduction}
The human visual system (HVS)~\cite{kruger2012deep} perceives visual contents by processing and integrating manifold information into abstract high-level concepts (\textit{e.g.}, structure, texture, semantics), which form the basis for subsequent cognitive process~\cite{zhang2020connecting}. 
From the perspective of machine vision, high-level visual concepts also play a more important role in the practical applications than signal-level pixels.
Existing compression methods, including traditional block-based (\textit{e.g.},~JPEG~\cite{pennebaker1992jpeg} and HEVC~\cite{sullivan2012overview}) and deep learning based methods~\cite{balle2019end,minnen2018joint}, mainly focus on the modeling and removal of signal-level redundancy, while the potential and versatility of accomplishing compression task through decomposing visual data into compact conceptual components still lack further exploration.
Following the insight of Marr~\cite{marr1982computational} and Guo~\textit{et al.}~\cite{guo2007primal}, visual objects usually appear as structures and textures.
As two primal visual components, structure and texture not only play a deterministic role in visual content synthesis, but also are critical visual feature descriptors in various analysis-based tasks. 
In particular, due to the lack of explicit modeling for structure, existing compression methods often introduce severe distortions in decoded images under low bit-rate scenarios, such as ringing/blocking artifacts and blurring edges, which not only degrade human perception, but also hamper the performance of visual analysis tasks. 
Moreover, the encoded bitstreams are less correlated to visual concepts, resulting in the difficulty of utilizing the encoded information for subsequent manipulation tasks, such as image synthesis, shape modification and content recreation.

Deep generative models, such as variational auto-encoders (VAE)~\cite{kingma2014auto} and generative adversarial networks (GANs)~\cite{goodfellow2014generative}, have offered a new approach for conceptualizing images with compact latent representations, where the decoding process is implemented in a generative fashion. 
However, existing researches on conceptual compression~\cite{gregor2016towards}\cite{santurkar2018generative} often attempt to capture image contents with a single latent vector, with different conceptual components entangled together. In consequence, the encoded conceptual representation is less interpretable and editable, limiting
its potential towards downstream image processing and machine vision tasks~\cite{gao2021digital}, such as target-guided content manipulation~\cite{wang2018high}. Apparently, it is desirable to develop visual components disentangled representations for a clearer understanding and more flexible control of image contents.

\begin{figure*}[t]
    \centering
    \vspace{-6mm}
    \includegraphics[width=1.0\linewidth]{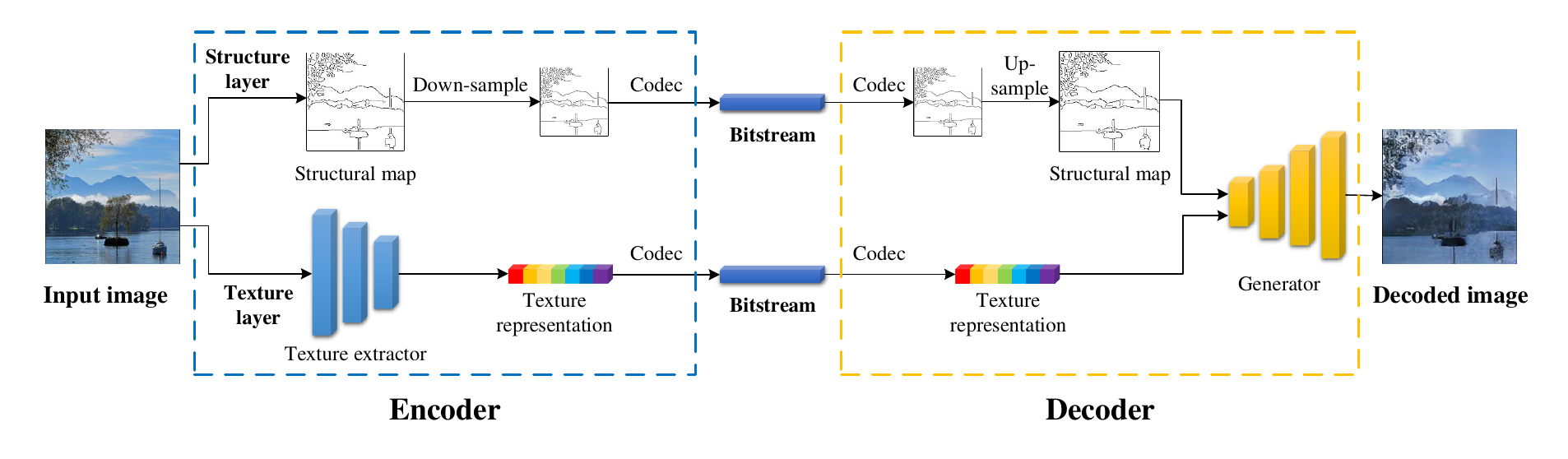}
    \vspace{-6mm}
    \caption{\textbf{Overview of proposed conceptual compression framework.} The framework consists of a structure layer and a texture layer. On the encoder side, structural maps and texture representations are extracted from input images using edge detector and texture extractor (VAE) respectively. Both two layers are compressed into bitstreams and reconstructed with specific codecs individually. A proposed generator integrates structure and texture to synthesize target images on the decoder side.}
    \label{fig:framework}
    \vspace{-6mm}
\end{figure*}

In this work, we propose a novel conceptual compression framework. In this framework, images are encoded into compact structure and texture representations and decoded in a deep synthesis fashion to achieve better visual reconstruction quality, flexible content manipulation and potential support for various analysis tasks. 
More specifically, we propose to process images by a dual-layered model consisting of two complementary visual features: 1) structure layer represented by edges since edges depict key structure information of images, and 2) texture layer which represents non-structural content including texture, color, luminance. 
As shown in Fig.~\ref{fig:framework}, at the encoder side, the structural maps are extracted first and then compressed to bitstream. 
Without complex texture, structural maps are more sparse and compression-friendly.
Meanwhile, the highly compact deep texture representations are extracted from input images with the variational auto-encoder in the form of low-dimensional latent variables and then scaled, quantized and entropy coded, leading to a bits-saving stream.
At the decoder side, a hierachical fusion GAN (HF-GAN) is proposed to integrate texture layer and structure layer to synthesize images after reconstructing texture representations and structural maps.
Apart from the image compression task, the texture pattern and synthesis paradigm are well jointly learned due to the fact deep generative models tend to capture unique statistical features. 
Thus, compared to encoding image into both a raw data layer and a specific feature layer separately~\cite{akbari2019dsslic,ma2018joint},
the proposed framework realizes the unification of visual features and basis data identity in the cross-modality sense, satisfying the demand of machine and human vision with one unified conceptual bitstream.  

Our methods are evaluated upon diverse images of fashion items~\cite{yu2014fine,zhu2016generative}, faces~\cite{karras2018progressive} and natural scenes, achieving significant perceptual gain against traditional block-based and deep learning-based end-to-end image compression frameworks. 
Furthermore, the versatility of the proposed conceptual coding scheme are verified upon a wide range of image processing tasks, including content manipulation, texture synthesis and face detection.

Our contributions can be summarized as follows:
\begin{itemize}
    \item We propose a novel compression framework which encodes by abstracting visual data into compact structure and texture representations and decodes by deep synthesis processing to produce cross-modal unification of visual features and basis data.
    \item We propose to realize conceptual compression via extracting sparse structural maps and deep texture representations, and an HF-GAN is proposed as decoder to reconstruct images of high visual quality from texture and structure.  
    \item The superiority of the proposed framework in image compression and vision tasks is justified and thoroughly analyzed via extensive experiments.
\end{itemize}

The rest of the paper is organized as follows. The related work and technologies are introduced in Section \ref{related work}. The proposed framework and networks are presented in Section \ref{proposed method}. Section \ref{experients} shows the detailed experimental results and analyses, and Section~\ref{conclusion} concludes the paper and discuss directions for future research.

\section{Related works}
\label{related work}
\subsection{Traditional Image Compression}
Traditional image compression technologies have played a fundamental role in visual communication and image processing, bringing up a series of crafted image codecs, such as JPEG~\cite{pennebaker1992jpeg}, JPEG2000~\cite{rabbani2002jpeg2000} and H.265/HEVC-based BPG~\cite{bellard2015bpg}.
JPEG is the most popular and widely used compression standard, which integrates well-known technologies including block-partition, discrete cosine transform (DCT), quantization  and entropy coding.
JPEG2000~\cite{rabbani2002jpeg2000} applies discrete wavelet transform (DWT) technology instead of DCT, achieving higher compression ratio and allowing various editing or processing applications.
After decades of development, a series of block-based hybrid prediction/transform coding standards (\textit{e.g.}, MPEG-4 AVC/H.264~\cite{wiegand2003overview}, AVS~\cite{gao2014overview,AVS3} and HEVC~\cite{sullivan2012overview}), are built to significantly improve the coding efficiency by reducing pixel-level redundancy.
Different from JPEG, HEVC utilizes more intra prediction modes from neighboring reconstructed blocks in spatial domain to remove redundancy. 

\begin{figure*}[t]
    \centering
    \vspace{-6mm}
    \includegraphics[width=0.95\linewidth]{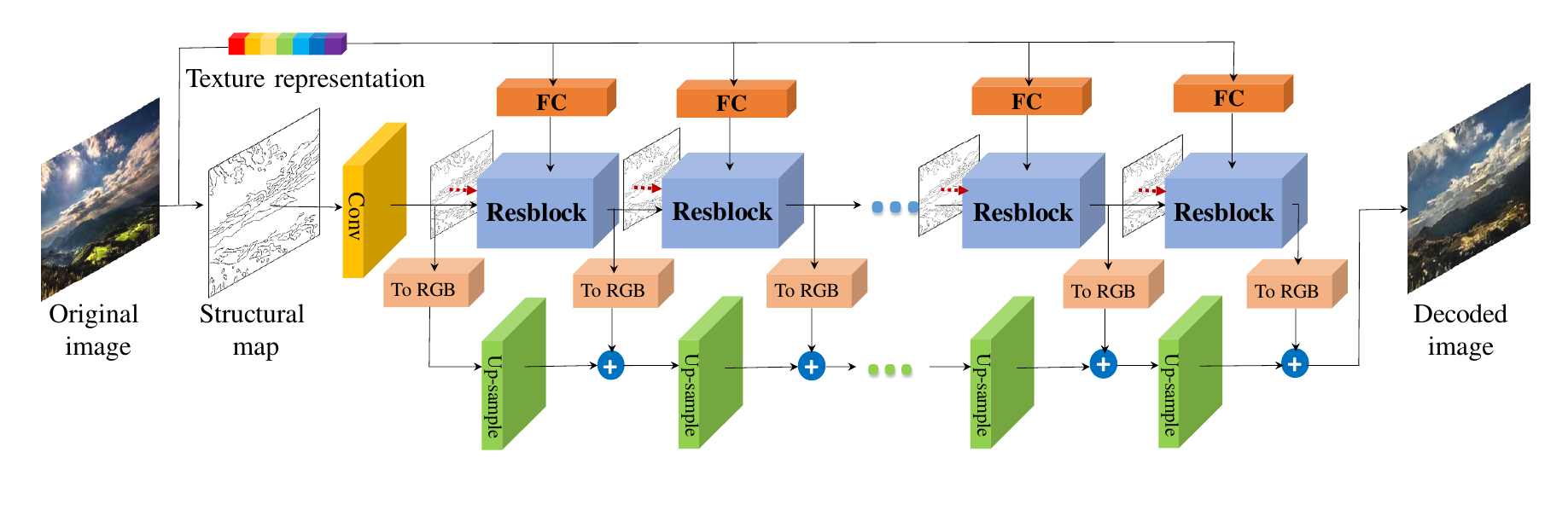}
    \vspace{-6mm}
    \caption{\textbf{Architecture of hierachical fusion generator networks.} Generator consists of fully connected layers (FC), Residual blocks (Resblock), RGB transformation module (To RGB), upsampling and cumulatively summing module. Structural maps are resized and concatenated to feature maps as input for corresponding Resblock.}
    \label{fig:networks}
    \vspace{-6mm}
\end{figure*}
\subsection{Deep Learning-based Image Compression}
Recent years have witnessed a surge of interest in learning-based image compression, which benefits from large-scale data, effective network architectures, end-to-end optimization, and other advanced techniques such as generative models and unsupervised learning. 
Typical deep image compression systems usually adopt an end-to-end framework~\cite{ma2019image} to model the pixel distribution and jointly optimize rate-distortion performance with reconstruction tasks, outperforming popular block-based codecs, such as JPEG, JPEG2000 and HEVC. 
Pixel probability modeling~\cite{salimans2017pixelcnn++} and auto-encoder~\cite{balle2019end,minnen2018joint} are two main approaches in deep learning based image coding schemes, as pointed in~\cite{liu2020deep}.
Meanwhile, multiple deep structures, such as CNN~\cite{balle2019end}, RNN~\cite{toderici2015variable} and GAN~\cite{santurkar2018generative,agustsson2019generative}, are utilized to explore efficient compression architectures.
Despite the constant improvement of compression performance, most deep learning-based end-to-end compression algorithms mainly aim to model signal-level correlations, while the visual concepts are less explored.
In addition to rate-distortion optimization, subsequent analysis tasks, such as image retrieval~\cite{zhang2017deep} and semantic analysis~\cite{luo2018deepsic,torfason2018towards}, have also been incorporated into learning-based compression frameworks to encourage analysis-friendly signal representation. 
However, extra task-related networks are adopted in their frameworks for joint image analysis and compression, potentially limiting the application to specific tasks. 
\subsection{Conceptual Compression}
Conceptual compression~\cite{gregor2016towards, chang2019layered, chang2021thousand} aims to encode images into compact, high-level interpretable representations for reconstruction, allowing a more efficient and analysis-friendly compression architecture. 
Gregor \textit{et al.}~\cite{gregor2016towards} introduce convolutional Deep Recurrent Attentive Writer (DRAW), which extends VAE by using RNNs as encoder and decoder, to transform an image into a series of increasingly detailed representations. 
However, the models in~\cite{gregor2016towards} are only verified on datasets of small resolutions.
Hu \textit{et al.}~\cite{hu2020towards} compress images into compact structure and color representations with generative models, where the representations are characterized with edge maps and reference sample pixels near edges specifically.
Despite that the structure representation can be well analyzed, the support for vision tasks, such as image synthesis and modification, are limited by pixel-level color representations. 
The semantic guidance is used to assist in reconstructing images from compact compressed visual data in~\cite{agustsson2019generative,akbari2019dsslic}, where the main data stream is still signal-oriented though.

\section{Conceptual image compression framework}
\label{proposed method}
Our goal is to explore the potential of performing compression via decomposing visual data into compact visual components and develop a set of feasible examplar scheme. 
In particular, we propose to compress images by extracting the compact representations of two primal complementary visual features, structure and texture.
In the proposed framework, the images ${I}$ are represented with compact texture representations $I^t$ and sparse structural maps $I^s$,~\textit{i.e.}, $I \to I^s + I^t$.
As shown in Fig.~\ref{fig:framework}, images are decomposed into texture layer and structure layer and compressed separately on the encoder side. 
We obtain the sparse structure maps $I^s$ with the edges extractor $\mathit{Enc^s}$ (Section~\ref{structure layer compression}) and low-dimensional texture representations $I^t$ with variational texture auto-encoder $\mathit{{Enc}^t}$ (Section~\ref{texture layer compression}), \textit{i.e.}, $I^s=Enc^s(I)$, $I^t=Enc^t(I)$. 
Both texture representations and sparse structural maps are further compressed to bitstreams for transmission using existing standard codec. 
On the decoder side, a hierarchical fusion generator $Gen$ is designed to reconstruct the high quality images $\hat{I}$ from decoded components, texture $\hat{I}^t$ and structure $\hat{I}^s$,~\textit{i.e.}, $\hat{I}=Gen(\hat{I}^t,\hat{I}^s)$ (Section~\ref{image synthesis}).
To achieve reconstruction of high visual quality and fidelity, we take advantage of both adversarial training and multi-perspective distortion measurement to supervise the reconstruction process.
In particular, we also propose a latent regression loss to better learn the paradigm of texture extraction and synthesis (Section~\ref{loss objectives}).
\subsection{Structure Layer Compression}
\label{structure layer compression}
Considering that edges are one of the most sparse and abstract image representations and can depict the key structure information of images, edge maps are extracted as the structure layer representation via widely used edge detection methods, such as holistically-nested edge detection (HED)~\cite{xie2015holistically} and Canny edge detection~\cite{ding2001canny}.
On account of the sparsity and binarization of structural maps, we employ the Lanczos downsampling algorithm to further reduce data volume of structure layer with 4-scale.
Furthermore, screen content coding~\cite{6783726} is adopted to compress the structure maps into bitstreams since this framework has strong capability of compressing images with abundant sharp edges.

At the decoder side, the process is reversed to recover the structural maps before synthesizing images.
To regain the structural maps of original resolution, we upsample the decoded low resolution structure maps using one of the state-of-the-art super-resolution methods, deep back-projection networks (DBPN)~\cite{haris2018deep}.
The DBPN model is optimized with the mean square error (MSE) in~\cite{haris2018deep} which is not sensitive to the fluctuation of sparse binary data, leading to the significant distortion in reconstructed edges. 
To improve the restoration performance on the datasets which are characterized by sparse and binary edges, the binary cross entropy (BCE) is employed to replace the MSE loss for the super resolution model training,
\begin{equation}
\begin{split}
\mathcal{L}_{BCE}=&-\sum_u \sum_v [I_{(u,v)}^s\log (\hat{I}^s_{(u,v)})\\ &+(1-I_{(u,v)}^s)\log(1-\hat{I}^s_{(u,v)})],
\end{split}
\end{equation}
where $I_{(u,v)}^s$ represents the true pixel value of structural map $I^s$ at position $(u,v)$, while $\hat{I}^s_{(u,v)}$ represents the predicted value of upsampled structural map $\hat{I}^s$ at position $(u,v)$.
We adopt the pre-trained DBPN model for high efficiency.
\subsection{Texture Layer Compression}
\label{texture layer compression}
As shown in Fig.~\ref{fig:framework}, the compact texture representations are extracted with a neural network-based encoder and integrated with the reconstructed structural maps to synthesize target images by an elaborate generator.
The encoder $Enc^t$ and generator $Gen$ are jointly trained in an end-to-end manner.
In this section, we will introduce the process of texture extraction and compression.

The image texture extractor is designed based on variational auto-encoder for conceptual level information extraction~\cite{gregor2016towards}.
In particular, the encoder $Enc^t$ is constructed with several residual blocks~\cite{he2016deep} and convolutional layers to model the input image $I$ into a posterior multivariate Gaussian distribution $q_{\phi}(I^t|I)$. 
Then the representation $I^t$ is generated by sampling from the posterior distribution $q_{\phi}(I^t|I)$ using re-parameterization~\cite{kingma2014auto} method, \textit{i.e.}, $I^t \sim q_{\phi}(I^t|I)$.

Subsequently, the extracted texture discrete representation $I^t$ is further compressed through scalar quantization and entropy coding. We follow the quantization method defined in HEVC~\cite{sullivan2012overview}, where the quantization step $Q_{step}$ is determined by the quantization parameter ($QP$) and we take the factor $s=2^{10}$ empirically:
\begin{equation}
    Q_{step}=\frac{2^\frac{QP-4}{6}}{s}=2^{\frac{QP-4}{6}-10}.
    \label{quantization}
\end{equation}
The quantization is performed by,
\begin{equation}
    I^{qt}=floor\left(\frac{I^t}{Q_{step}}\right),
    \label{floor}
\end{equation}
where representation $I^t$ is quantized to $I^{qt}$ and $floor(\cdot)$ represents the truncation operation for designated binary digit precision.
Finally the quantized texture representations are encoded with Arithmetic codec~\cite{witten1987arithmetic} for transmission. 
The decoded texture representations for image synthesis can be denoted by $\hat{I}^t$, where $\hat{I}^t=I^{qt}*Q_{step}$.
\begin{figure}[t]
    \centering
    \vspace{-4mm}
    \includegraphics[width=1.0\linewidth]{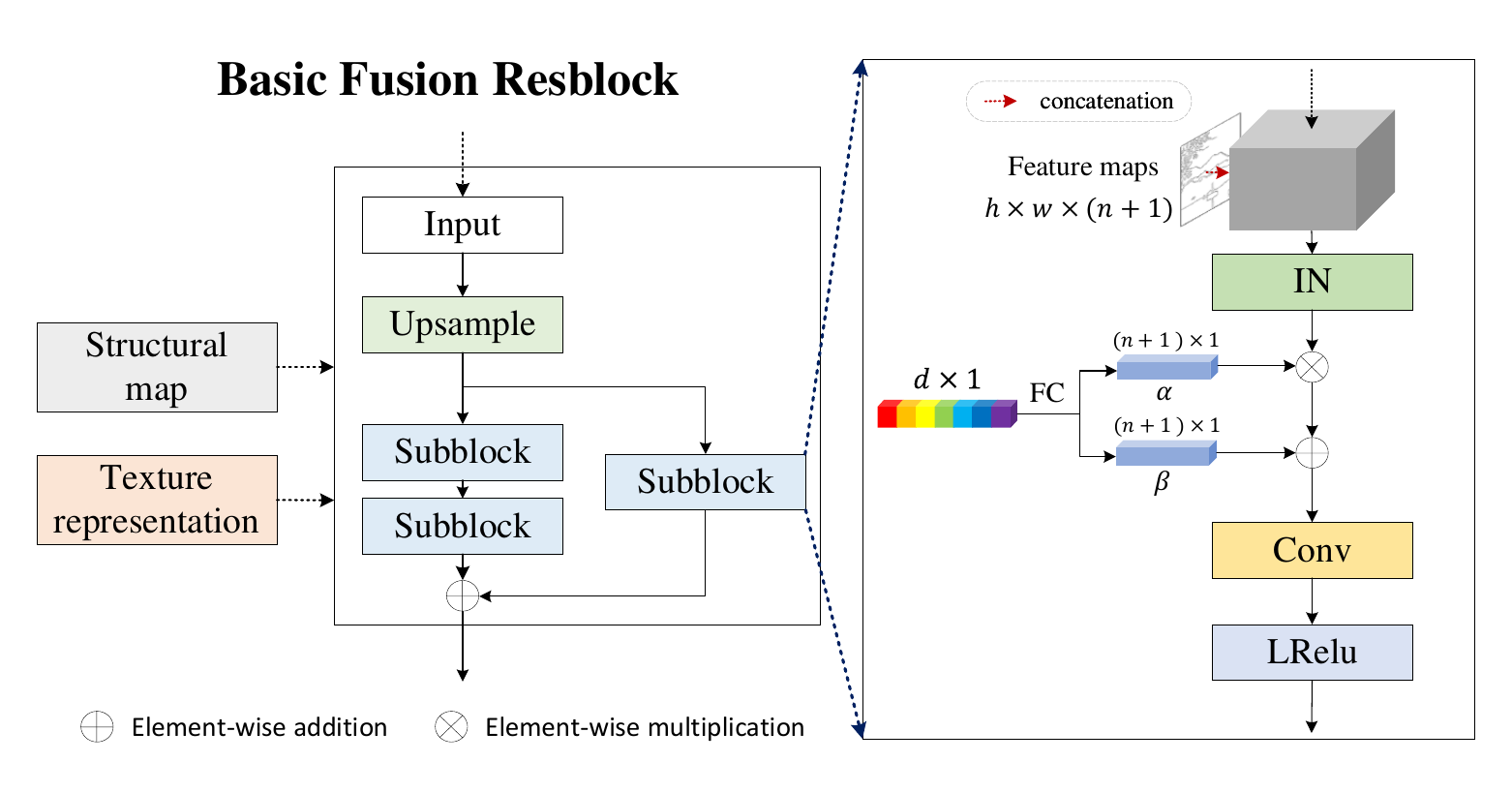}
    \vspace{-8mm}
    \caption{\textbf{Basic fusion residual block in generator.} Resblock input ($h \times w \times (n+1)$) consists of feature maps ($h \times w \times n$) and a structural map ($h \times w \times 1$). Texture representation ($d \times 1$) is transformed into two groups of affined parameters: $\alpha$ ($(n+1)\times 1$) and $\beta$ ($(n+1) \times 1$). Structural maps and texture representations are fused through parameterized normalization in each subblock.}
    \label{fig:resblock}
    \vspace{-5mm}
\end{figure}

\subsection{Image Synthesis}
\label{image synthesis}
Motivated by the strong capability of the instance normalization in style transfer~\cite{karras2020analyzing}, we adopt the adaptive instance normalization~\cite{huang2017arbitrary} to integrate texture layer and structure layer. 
Specifically, a group of modulation parameters, including mean and variance, are learned from texture representations.
The feature statistics of the feature maps are transferred with learned modulation parameters during adaptive affine transformations, where feature maps contain both structure information and previously generated texture information in forward propagation as shown in Fig.~\ref{fig:resblock}.
The particular channel-wise modulation parameters are learned corresponding to each intermediate feature map. 
In this manner, the structure and texture are fused and integrated gradually through convolutional operation and affine transformation. 
Additionally, inspired by impressive generation capability of progressive growing GAN~\cite{karras2018progressive}, we devise a generator which progressively increases the resolution of synthesized feature maps as shown in Fig.~\ref{fig:networks}.  
The generator consists of a pile of residual blocks as basic units and takes advantage of skip connections and hierarchical fusion. 
Each residual block includes three fully convolutional layers followed by adaptive instance normalization operation. The structural maps are concatenated to the feature maps as new input for each residual block. 
Finally, the target images are obtained by upsampling and summing the contributions of RGB outputs corresponding to different resolutions as~\cite{karras2020analyzing}.
The formulation of the hierarchical fusion process is presented as follows. 
Given the input image $I \subseteq \mathbb{R}^{H\times W \times 3}$, we can obtain the structure map $I^s \subseteq \mathbb{R}^{H\times W \times 3}$ and the texture representation $I^t \subseteq \mathbb{R}^{d \times 1} $, where $H,W$ represent the size of input image and $d$ denotes the dimension of texture representation. 
Let the initial block of the generator function $\mathcal{G}_0$ be defined as $\mathcal{G}_0:I^s \mapsto A_0$ such that $A_0 \subseteq \mathbb{R}^{2\times 2 \times c_0}$ and $c_0$ denotes the number of feature channels. 
We define the intermediate output of the $i$-th residual block as $A_i$, and $i \in \mathbb{N}$. The structure map $I^s$ is resized to $I^s_i$ which has the same size as $A_i$. Thus, the input of each residual block is defined as:
\begin{equation}
    \bar{A}_i=[A_i;I^s_i],
\end{equation}
where [;] is a channel-wise concatenation operation. Let $\mathcal{G}_i$ be a generic function where $\mathcal{G}_0$ acts as the initial convolutional layer and $\mathcal{G}_i$ acts as the basic Resblock: $\mathcal{G}_i: \bar{A}_{i-1} \mapsto A_i$, where $A_i \in \mathbb{R}^{2^{i+1}\times 2^{i+1}\times c_i}$ and $c_i$ is the number of channels in the intermediate activations of the $i$-th generator block. Then $A_k$ can be obtained by the general formula composed with a sequence of $\mathcal{G}$ functions, where $k \in \mathbb{N}$:
\begin{equation}
\begin{aligned}
    A_k = & \mathcal{G}_k(\cdot,I^s_k) \circ \mathcal{G}_{k-1}(\cdot,I^s_{k-1}) \circ \cdots \mathcal{G}_i(\cdot,I^s_i)\\
    & \circ \cdots \mathcal{G}_1(\cdot,I^s_1) \circ \mathcal{G}_0(I^s).
\end{aligned}
\end{equation}

The symbol $\circ$ in Eq (5) is defined as the function composition operation. In particular, $A_k$ can be unfolded with composition function as following:
\begin{equation}
\begin{aligned}
    A_k = g_k(g_{k-1}(\cdots g_i(\cdots g_1(g_0(I^s),I^s_1) \cdots,I^s_i) \cdots,I^s_{k-1}),I^s_k)
\end{aligned}
\end{equation}

We employ skip connections and multi-scale fusion to improve reconstruction quality of the generator. $up^{(2)}$ is defined as the upsampling function by 2-scale and $conv^{(3\times 3)}$ simply serves as a $3 \times3 $ convolution function which converts the output feature map $A_i$ of each block into RGB images: $conv^{(3\times 3)}_i:A_i \mapsto B_i$, where $B_i\subseteq \mathbb{R}^{2^{i+1}\times 2^{i+1}\times 3}$. Hence, the RGB output of initial convolutional layer $x_0$ is:
\begin{equation}
    x_0= conv^{(3\times 3)}_0(A_0)=conv^{(3\times 3)}_0(\mathcal{G}_0(I^s)).
\end{equation}
The RGB output of the $i$-th intermediate block can be computed with the recurrence formula as following:
\begin{equation}
    x_i=conv^{(3\times 3)}_i(A_i)+up^{(2)}(x_{i-1}), i\geq1.
\end{equation}
The network starts incremental reconstruction process from the resolution of $4\times 4$ at the first Resblock, conducts $2\times$ up-sampling and fusion affine transformation in each Resblock, and obtains the target resolution output at the final Resblock.  
The selection of the number of Resblocks in hierarchical fusion generator network depends on the target image resolution of the dataset.
For the target synthetic resolution $n\times n$, the number of Resblocks is set to $\left \lceil \log_{2}{n - 1} \right \rceil$.
Therefore, the proposed generator consists of seven Resblocks for the target resolution $256\times 256$.

\begin{figure*}[t]
    \centering
    \vspace{-11mm}
    \includegraphics[width=0.9\linewidth]{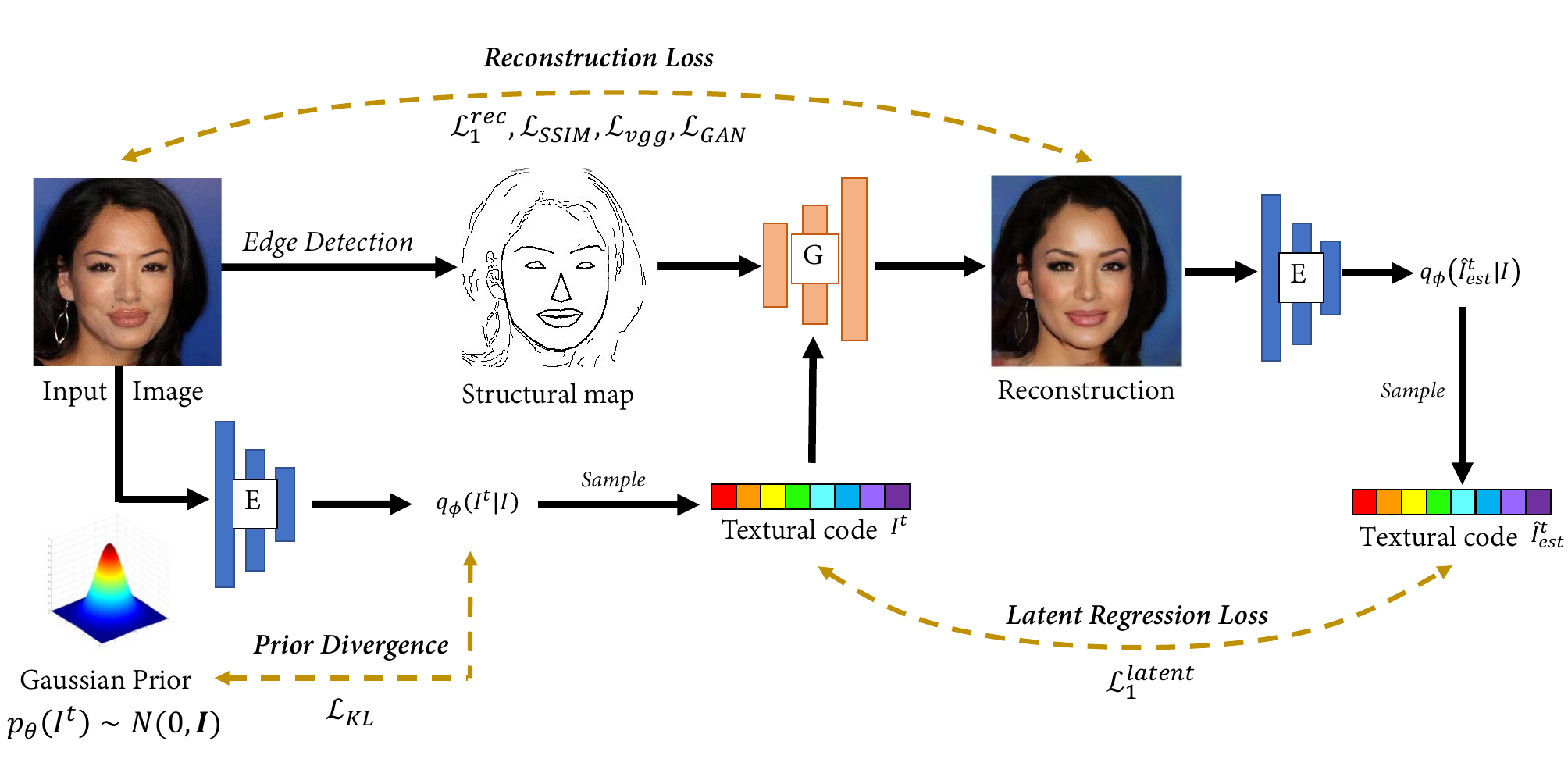}
    \vspace{-4mm}
    \caption{\textbf{Loss functions of end-to-end VAE-HFGAN.} During training, various losses are adopted: for adversarial training, a discriminator is adopted to discriminate real labels and fake labels; to reduce the distortion of reconstructed images, we apply self reconstruction $L_1$ loss, structure similarity loss $SSIM$ and perceptual $L_1$ loss which measure the distance between features encoded from \textit{VGG} networks; to align texture representations with a prior Gaussian distribution, a $KL$ loss is incorporated; to enforce the connection between paired texture representations and images, the latent regression $L_1$ loss is employed to measure the difference between texture representations extracted from original image and reconstructed image. }
    \label{fig:loss}
    \vspace{-5mm}
\end{figure*}
In the basic residual block of the generator shown in Fig.~\ref{fig:resblock}, there are three convolutional layers whose scale and bias are modulated by three groups of parameters learned from texture representations by three fully connected layers respectively. 
The feature maps at the site of $j$-th convolutional layer of $i$-th basic block is denoted by $A_{ij}$. Receieving the texture representation $I^t$, a fully connected function $f_{ij}:I^t \mapsto {\alpha_{ij},\beta_{ij}}$ first maps $I^t$ into affine parameters, where $\alpha_{ij},\beta_{ij} \subseteq \mathbb{R}^{1\times 1 \times c_i}$. Thus, we adaptively normalize the previous activations $A_{ij}$ with the learnable texture affine parameters:
\begin{equation}
    AdaIN(A_{ij},I^t)=\alpha_{ij} \frac{A_{ij}-\mu(A_{ij})}{\sigma(A_{ij})}+\beta_{ij},
\end{equation}
where the normalized feature maps are scaled with $\alpha_{ij}$, and shifted with $\beta_{ij}$ in the channel-wise manner, and $\mu(A_{ij})$ and $\sigma(A_{ij})$ are the means and standard deviations of the activations in $A_{ij}$.
By transferring the feature statistics from texture representations and concatenating the structural maps with intermediate generated content, structure and texture are effectively integrated to reconstruct images of which texture are progressively enriched.
Moreover, during the training of reconstruction task, the mapping and synthesis paradigm between deep latent space and spatially-aware texture content are jointly well learned and stored as deep prior in the designed generative model, allowing the flexible examplar-based image content manipulation in the compressed domain.
\subsection{Loss Objectives}
\label{loss objectives}
The proposed texture encoder and the image generator are jointly trained in an end-to-end manner with a multi-scale discriminator. 
The image compression and reconstruction tasks are mainly optimized on three type of losses as shown in Fig.~\ref{fig:loss}: \textit{\textbf{reconstruction loss}} which aims to improve reconstruction visual quality and fidelity, \textit{\textbf{prior divergence}} which provides a prior distribution instruction for deep texture representation, and the proposed \textit{\textbf{latent regression loss}} which constraints the mapping between deep latent space and synthsis texture content.

Compared to typical learned lossy compression methods~\cite{balle2019end,minnen2018joint} which only optimize image reconstruction quality with pixel-wise similarity metric, we introduce diverse distortion metrics to supervise reconstruction task from low level pixel-wise loss to high level perceptual loss and adversarial loss. The~\textbf{reconstruction loss} is introduced as follows:
\begin{itemize}
    \item Self-reconstruction loss: the pixel-wise loss $\mathcal{L}{^{rec}_{1}}$ is imposed to force the visual components to complete reconstruct the original images: 
    \begin{equation}
    \mathcal{L}{^{rec}_{1}} = \mathbb{E}_{I\sim p(I), {I^t} \sim p(I^t)}\left\|{I-Gen(\hat{I}^t,\hat{I}^s))} \right \|_1.
    \end{equation}
    \item Structure similarity loss: the SSIM~\cite{wang2004image} loss is incorporated to supervise the optimization process of improving structural fidelity: 
    \begin{equation}
    \mathcal{L}_{SSIM}=\mathbf{SSIM}(I,Gen(\hat{I}^t, \hat{I}^s)).
    \end{equation}
    \item Perceptual loss: we use the perceptual loss $\mathcal{L}_{vgg}$ in~\cite{johnson2016perceptual} to encourage the perceptual fidelity through deep feature matching from pre-trained VGG-16 netoworks~\cite{simonyan2014very}.
    \item Adversarial loss: we apply the discriminator $Dis$ and follow the variant of conditional adversarial training scheme~\cite{mao2017least} to encourage the visual realism of reconstructed images:
    \begin{multline}
         \mathcal{L}{^{GAN}_{G}} = \frac{1}{2}\mathbb{E}_{I\sim p(I)}||1-Dis(I,I^s)||_2 \\ +
         \frac{1}{2}\mathbb{E}_{I\sim p(I), I^t \sim p(I^t))}||Dis(Gen(\hat{I}^t, \hat{I}^{s}),I^s)||_2,
    \end{multline}
    \begin{multline}
        \mathcal{L}^{GAN}_{D} = -\mathbb{E}_{I\sim p(I), I^t \sim p(I^t))} \\ || Dis(Gen(\hat{I}^t, \hat{I}^{s}),I^s)||_2.
    \end{multline}
\end{itemize}

Additionally, to obtain meaningful texture representations by stochastic sampling and benefit entropy coding process, we introduce the \textbf{prior divergence} to enforce the distribution of extracted texture representation to be close to the prior Gaussian distribution,~\textit{i.e.} $p_{\theta}(I^t)\sim\mathcal{N}(0,\mathbf{I})$ and $\mathbf{I}$ denotes the identity matrix:
\begin{equation}
\mathcal{L}_{KL} = \mathbb{E}_{I\sim p(I)}[\mathcal{D}_{KL}(q_{\phi}(I^t|I)\|p_{\theta}(I^t))].
\end{equation}

Furthermore, we propose a \textbf{latent regression loss} to further constrain the bidirectional mapping between the learned texture representation ${I^t}$ and the synthesis texture content $\hat{I}$.
More specifically, the texture encoder $Enc^t$ is utilized to analyze and extract the texture representation $\hat{I}^t_{est}$ of the generated image $\hat{I}$.
Given the fact that the reconstructed images $\hat{I}$ are trained to maintain texture fidelity, the texture representations which are extracted from original images and reconstructed images should be encouraged to be consistent,~\textit{i.e.}~$I^t = Enc^t(I) \approx \hat{I}^t_{est} = Enc^t(\hat{I})$.
Thus, the latent regression loss $\mathcal{L}{^{latent}_1}$ is proposed to minimize the $L_1$ distance between two texture representations as follows,
\begin{equation}
    \mathcal{L}{^{latent}_1}=E{_{I^t \sim p(I^t),\hat{I}^t_{est} \sim p({\hat{I}^t_{est})}}} \|{I^t-\hat{I}^t_{est}} \|_1.
\label{con:latent}
\end{equation}
The unique association between specific representation and corresponding texture are further reinforced by the latent regression loss, which not only helps learn a continuous latent space for texture expression in the data-driven manner, but also benefits the learning of synthesis paradigm. 
Above all, the final training objective is shown as following:
\begin{multline}
\label{loss_G}
    \mathcal{L}_{G,D,E}= \lambda_{GAN} \mathcal{L}{^{GAN}_{G}}+\lambda_{ rec}\mathcal{L}_{1}^{rec}+
    \lambda_{vgg}\mathcal{L}_{vgg}
    \\+\lambda_{SSIM}\mathcal{L}_{SSIM}+ +\lambda_{KL}\mathcal{L}{_{KL}} +\lambda_{ latent}\mathcal{L}_{1}^{latent},
\end{multline}
where the hyper-parameters $\lambda_{name}$ weights each loss respectively.

\begin{figure*}[t]
    \centering
    \vspace{-6mm}
    \includegraphics[width=1.0\linewidth]{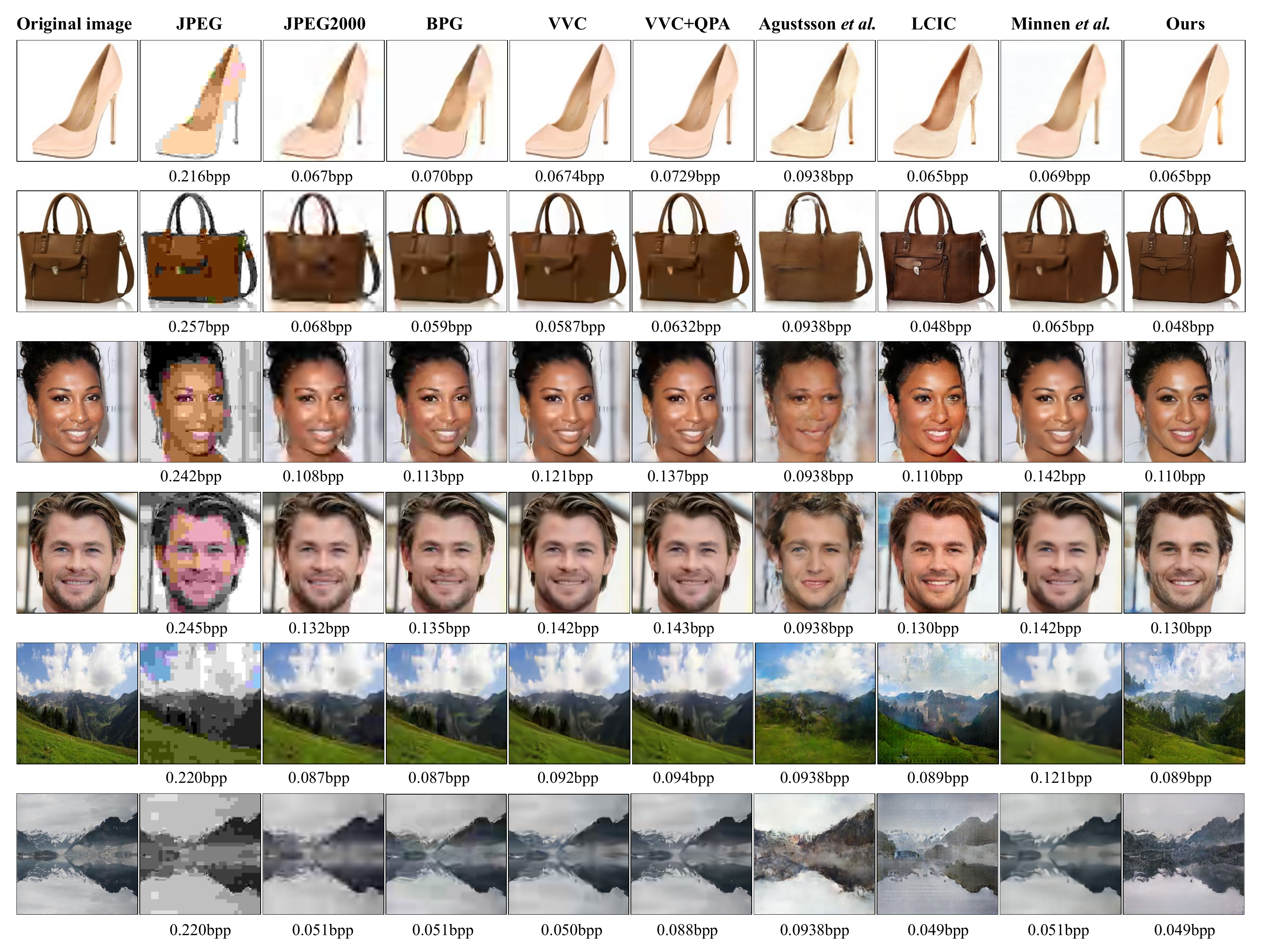}
    \caption{{\textbf{Qualitative comparisons with baselines under similar low bit-rate.} The \textbf{\textit{bpp}} of each image is reported under corresponding image. 
    }}
    \label{fig:qualitative}
    \vspace{-2mm}
\end{figure*}
\section{Experiments}
\label{experients}
\subsection{Implementation details}
We implement the proposed model using PyTorch and trained on two NVIDIA Tesla V100 GPUs. 
For training, we use the Adam optimizer with exponential decay rates $(\beta_1,\beta_2)=(0.5,0.999)$. 
The batch size is set to 16 and the learning rate is set to 0.0002.
The training procedure follows the Least Squares GANs (LSGANs)~\cite{mao2017least}.
We adopt the following hyper-parameters in all experiments for the training: $\lambda_{GAN}=1.0, \lambda_{rec}=10.0, \lambda_{SSIM}=0.25, \lambda_{vgg}=0.2, \lambda_{latent}=1.0, \lambda_{KL}=0.01$. 
As for the size of texture representations, we find that the capacity and capability of texture representations are growing as size increases. Thus, factored by texture complexity of specific datasets, the best sizes vary for different datasets. 
In our experiments, the dimension of texture representations is empirically set to $d = 64$ across all datasets for comparison.  
\subsection{Datasets}
We conduct experiments on several datasets including edges2shoes~\cite{yu2014fine}, edges2handbags~\cite{zhu2016generative}, CelebA-HQ~\cite{karras2018progressive}, and the multiple seasons dataset. All images are resized to $256\times 256$ in the experiments.

\textbf{Edges2shoes and edges2handbags.}
We combine these two datasets for training due to their high content similarity.
It contains $188,392$ paired training images and $400$ images for testing.
The images of edges are utilized for providing structural information.

\textbf{CelebA-HQ.} It includes $30000$ high-quality face images~\cite{karras2018progressive}.
We split $29800$ as training set and $200$ as testing set.
For structural maps, we employ facial landmark detection to obtain contours in the facial region, and the Canny edge detector~\cite{ding2001canny} to obtain structural edges in the background region.   

\textbf{Multiple seasons dataset. }
We collect $8000$ images from the Yosemite dataset~\cite{zhu2017unpaired} and the alps seasons dataset~\cite{anoosheh2018combogan}.
The dataset contains four seasons of images, and $400$ images are split for testing.
We use the Canny edge detector incorporated with the Gaussian blur algorithm to acquire the corresponding structural maps. 

\begin{figure*}[t]
    \centering
    \includegraphics[width=1.0\linewidth]{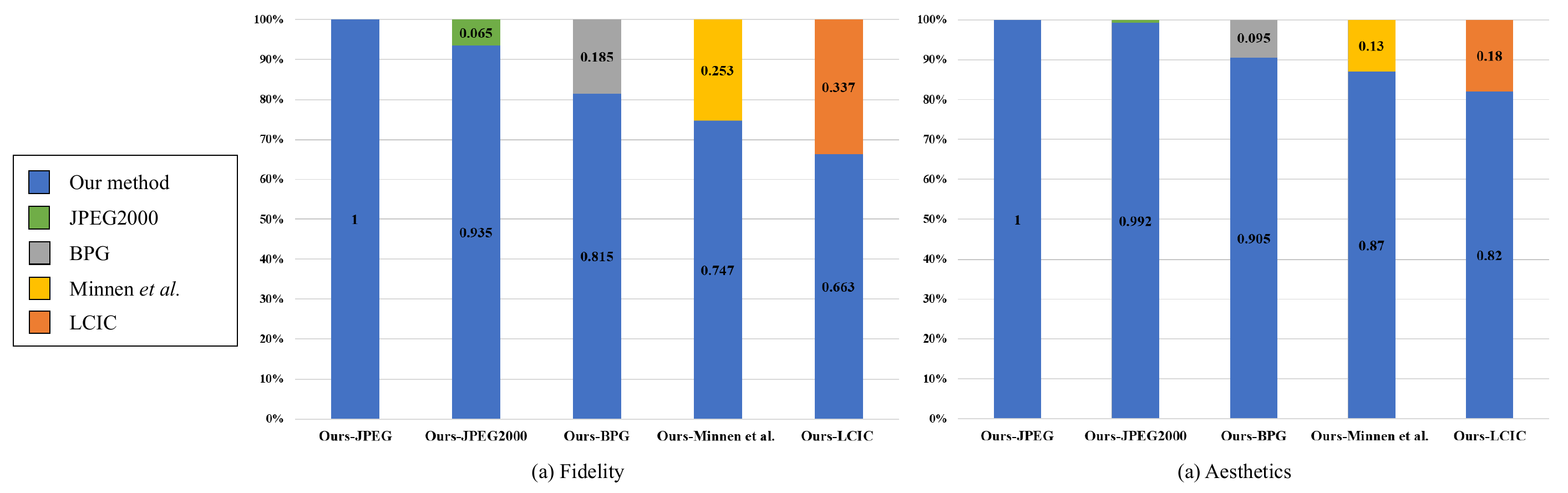}
    \vspace{-6mm}
    \caption{ {\textbf{Fidelity and aesthetics preference results.}} A user study is conducted, in which participants are asked to choose results which better match original images (Fidelity) and have better perceptual quality (Aesthetics) through pairwise comparisons. The numbers in histogram indicate the percentage of preference for corresponding comparison pair. The results show that the proposed method obtains the best preference ratio in each comparison pair for both fidelity and aesthetics.}
    \label{fig:user_study}
    \vspace{-4mm}
\end{figure*}

\begin{table*}[t]
    \centering
    \caption{{Quantitative results on edges2shoes$\&$handbags, CelebA-HQ and multiple seasons datasets. $\downarrow$ means lower score is better. Bold and Underline indicate the best and second best values, respectively.}}
    \begin{tabular}{lccccccccc}
    \toprule
          Datasets & \multicolumn{3}{c}{Edges2shoes$\&$handbags} & \multicolumn{3}{c}{CelebA-HQ} & \multicolumn{3}{c}{Multiple seasons} \\
         \hline
         & Average bitrate & LPIPS$\downarrow$ & DISTS$\downarrow$ 
         & Average bitrate & LPIPS$\downarrow$ & DISTS$\downarrow$ 
         & Average bitrate & LPIPS$\downarrow$ & DISTS$\downarrow$ \\
         \hline
         JPEG & 0.241 bpp & 0.274 & 0.326 & 0.239 bpp & 0.492 & 0.400 & 0.227 bpp & 0.517 & 0.520 \\
         JPEG2000 & 0.060 bpp & 0.309 & 0.265 & 0.122 bpp & 0.289 & \underline{0.241} & 0.076 bpp & 0.532 & 0.536 \\
         LCIC & 0.059 bpp & 0.308 & 0.232 & 0.074 bpp & 0.234 & 0.272 & 0.076 bpp & 0.467 & 0.447 \\
         BPG & {0.044} bpp & \textbf{0.122} & \textbf{0.159} & 0.075 bpp & 0.249 & 0.276 & 0.052 bpp & 0.367 & 0.442 \\
         VVC & 0.033 bpp & 0.151 & 0.191 & 0.072 bpp & 0.229 & 0.254 & 0.043 bpp & 0.365 & 0.440 \\
         VVC+QPA & 0.034 bpp & 0.150 & 0.189 & 0.074 bpp & \underline{0.224} & 0.250 & 0.045 bpp & 0.353 & 0.425 \\
         Minnen \textit{et al.}& 0.030 bpp & 0.180 & 0.221 & 0.076 bpp & 0.225 & 0.246 & {0.072} bpp & \underline{0.305} & \textbf{0.367} \\
         Proposed method & {0.031} bpp & \underline{0.148} & \underline{0.181} & 0.074 bpp & \textbf{0.194} & \textbf{0.221} & {0.043} bpp & \textbf{0.303} & \underline{0.389} \\ 
         \bottomrule
    \end{tabular}
    \label{tab:quatitative_results}
    \vspace{-4mm}
\end{table*}
\subsection{Compression Performance Comparison}
In this subsection, we conduct extensive qualitative and quantitative experiments to compare the compression performance with traditional and learning-based approaches. 
To compress structure and texture representation into bitstreams with high efficiency, we adopt different strategies for structure layer and texture layer compression.
In particular, the reference software of screen content coding~\cite{6783726} is employed to compress the structural maps. 
Moreover, texture representations are quantized with $QP=51$ (Eq.~\ref{quantization}) and truncated with $16$ bits range (Eq.~\ref{floor}).
Then, the quantized results are encoded with lossless Arithmetic codec~\cite{witten1987arithmetic}.
Texture representations take less than $10\%$ of the total bits cost and most of the bits originate from edge maps coding.
Thus, we adjust the QP of the codec of edge maps to obtain bitrate coding results at different bitrates for the proposed method.

\subsubsection{Baselines}
We compare the proposed method with traditional compression standards and learning-based compression methods on the test datasets.
For the traditional compression codecs, the widely-used compression standards JEPG and JPEG2000, HEVC-based image compression codec BPG, and the latest standard Versatile Video Coding  (VVC)~\cite{bross2021overview} are applied for comparisons.
Moreover, some prior works aim to optimize conventional codecs using perceptual metrics to improve the perceptual quality~\cite{wang2011ssim,helmrich2019perceptually,sun2019perceptual}.
We also provide a comparison with a perceptually optimized method based on VVC using quantization parameter adaptation (VVC$+$QPA)~\cite{helmrich2019perceptually}.
For deep learning-based image compression approaches, we compare with Minnen \textit{et al.}~\cite{minnen2018joint} which adopts end-to-end optimization, previous generative compression Agustsson \textit{et al.}~\cite{agustsson2019generative}, and an existing conceptual compression method (LCIC)~\cite{chang2019layered}. LCIC employs basic VAE-GAN architecture for encoder and decoder, where the generator applies the U-Net structure and combines edge maps and latent codes by direct concatenation as input.
The specific settings are detailed as follows:
\begin{itemize}
    \item \textbf{JPEG}: we use JPEG Encoder of Matlab with quality factor $QF = 1$, which obtains maximum compression ratio of JPEG.
    \item \textbf{JPEG2000}: the images are compressed by OpenJPEG platform with quality parameter aligned with the compression ratio of proposed method, ranging from $200$ to $500$.
    \item \textbf{BPG}: we use standard bpg codec\footnote{https://bellard.org/bpg} with $QP$ ranging from $40$ to $51$ for compression ratio alignment.
    {\item \textbf{VVC}: We adopt the intra mode in VVC reference software VTM $11.0$\footnote{https://vcgit.hhi.fraunhofer.de/jvet/VVCSoftware\_VTM} and present reconstruction results at a similar bitrate to ours for comparison.
    \item \textbf{VVC$+$QPA}: Perceptually optimized method~\cite{helmrich2019perceptually} has been adopted in the reference software of VVC. Thus, we conduct testings on VTM with \textit{QPA} macro turned on.
    \item \textbf{Agustsson \textit{et al.}~\cite{agustsson2019generative}}: We provide the experimental results of Agustsson \textit{et al.} using a reproduced implementation\footnote{https://github.com/Justin-Tan/generative-compression}. For fair comparisons, we have trained the models of generative compression without adding noise on the same datasets as ours with 120 epochs.
    \item \textbf{LCIC}: We trained the model of LCIC\footnote{https://github.com/changjianhui/LCIC-pytorch} on the same datasets as ours for fair comparisons. The dimensionality of texture latent codes is set to 64 to compare reconstruction quality at equal bitrate.
    }
    \item \textbf{Minnen \textit{et al.}~\cite{minnen2018joint}}: The deep compression network is trained exactly following the procedure in~\cite{minnen2018joint} with hyper prior. We set $\lambda=0.0025$ for obtaining average 0.16 bits per pixel (bpp), and $\lambda=0.005$ for average 0.2 bpp. 
\end{itemize}

\subsubsection{Qualitative Evaluations}
For the evaluation of visual reconstruction quality, we mainly focus on the global structural fidelity and aesthetic sensibility. 
Making comparisons at the same bitrate is difficult since most compression methods cannot generate a specified bitrate.
As such, we select comparison images with sizes that are as close as possible to our encoding bpp.
We present qualitative comparisons in Fig.~\ref{fig:qualitative}.
In particular, the decoded images of JPEG and JPEG 2000 show serious structure and color distortion and blocking artifacts under extreme low bit-rate scenarios resulting from the limitation of block-wise processing.
The results of BPG show more distortion and degradation in visual quality than which of ours, \textit{e.g.}, the face skin looks unnatural and partial key edges are missing in the second row. 
Meanwhile, the structure and texture are over smoothed and blurred in VVC and Minnen \textit{et al.}~\cite{minnen2018joint}, hence losing significant structural details and reducing visual reality compared to our method, \textit{e.g.}, the teeth and jawline in the third row of Fig.~\ref{fig:qualitative}. 
The perceptually optimized VVC scheme does not improve visual quality compared to VVC but with a slightly higher bitrate, indicating that such methods cannot improve perceptual quality of traditional codec well under extreme low-bitrate setting. 
The reproduced results of Agustsson \textit{et al.} exhibits severe distortion and degraded image quality compared to ours.
Regarding the existing conceptual compression method, LCIC achieves visually pleasing reconstruction on edge2shoes and edge2handbags datasets. 
Nonetheless, the synthesis images appear apparent visual distortion and artifacts on complex scenarios, \textit{e.g.}, the checkerboards artifacts on the decoded natural scene images.
Overall, the results demonstrate that our method can produce reconstructed images of high visual quality under extremely low compression bitrate and outperform the baselines in regard to visual fidelity and aesthetic sensibility. 

\begin{figure*}[t]
    \centering
    \vspace{-2mm}
    \includegraphics[width=1.0\linewidth]{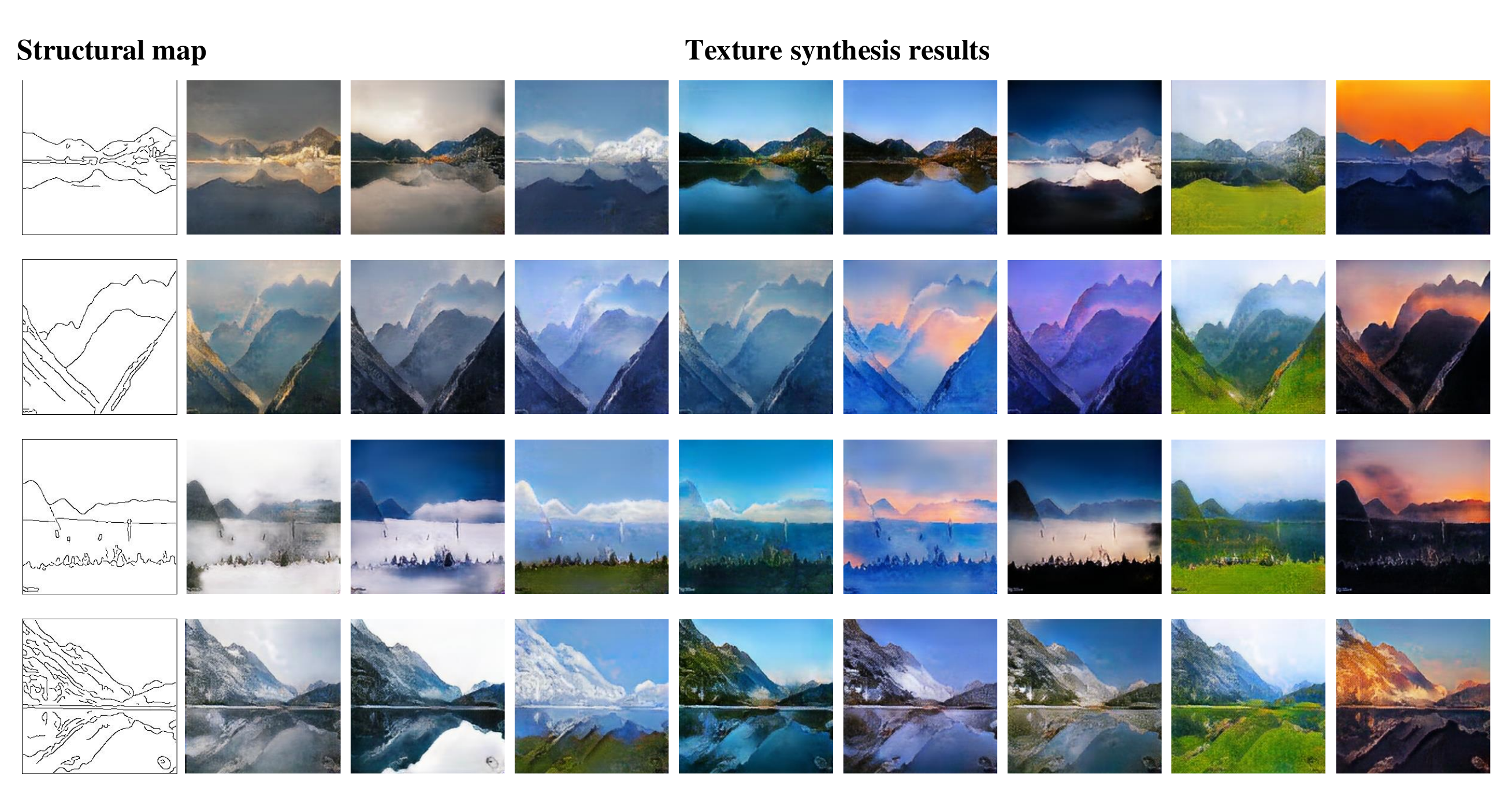}
    \caption{\textbf{Texture synthesis results.} With all accessible images serving as a texture source library, the proposed method is able to synthesize different texture by replacing the texture representation with ones encoded from images of desired texture styles, and generate creative and attractive images.}
    \label{fig:texture_transfer}
    \vspace{-4mm}
\end{figure*}
\subsubsection{Quantitative Evaluations}
Due to the ultimate arbiter of lossy compression remaining human evaluation, we choose LPIPS~\cite{zhang2018unreasonable}, DISTS~\cite{ding2020image} and user preference as perceptual quality metrics to evaluate the fidelity quantitatively, which prove to be highly correlate with human quality judgements~\cite{li2021quality} instead of only assessing signal fidelity.
Lower score of LPIPS and DISTS indicates higher visual fidelity of reconstructed images.
The average results of LPIPS and DISTS, and corresponding average bitrate over three datasets from different methods are shown in Table~\ref{tab:quatitative_results}.
It should be noted that we provide the available lowest bitrate of BPG on the edge2shoes and edge2handbags datasets, though still higher than which of our method.
It can be clearly seen that the proposed method almost excels all other comparison methods on DISTS and LPIPS under similar bitrate (except for BPG obtains better score of LPIPS and DISTS on edges2shoes and edges2handbags testing set, and Minnen \textit{et al.} obtains better DISTS score on multiple seasons testing set, while both of them take more bits for coding).

A user study of pairwise comparison is conducted to evaluate fidelity and aesthetics.
Given image pairs, users are asked to choose the one matching original image better (fidelity) in the first part of the survey, and the one of better visual quality (aesthetic) in the second part.
16 cases from test data of three datasets are selected to show for comparison.
A total of 46 subjects participate in the user study and a total of 2944 selections are tallied. 
User preference results are shown in Fig.~\ref{fig:user_study}. 
The proposed conceptual compression method obtains the best preference ratio in each comparison pair for both fidelity and aesthetics.
The quantitative results fully verify the superiority of our method in obtaining better visual quality and maintaining higher structure and texture fidelity.

\begin{figure}[!h]
    \centering
    \includegraphics[width=0.8\linewidth]{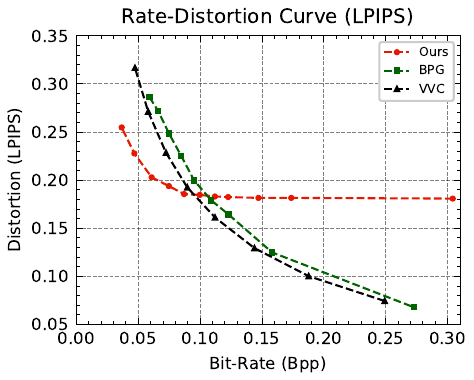}
    \caption{{\textbf{The rate-distortion curves.} The quantizer of the proposed method moves from lossless to coarse, while the corresponding compression performance of BPG and VVC are provided. The lower score of LPIPS demonstrates better perceptual quality. }}
    \label{fig:RD}
    \vspace{-4mm}
\end{figure}

{\subsection{Graceful Quality Degradation}
To have a better understanding of the performance comparison, we evaluate graceful quality degradation by varying bitrate for both the proposed method and conventional image codecs BPG and VVC.
We adjust the compression level of edge maps to obtain variable bitrate coding results for the proposed method and present the similar bitrates of BPG and VVC, shown in the rate-distortion curves of Fig.~\ref{fig:RD}.
Moving the quantizer from lossless to coarse, the bitrate of the proposed is mainly varied at the range of $(0, 0.3)$ bpp.
As shown in Fig.~\ref{fig:RD}, the proposed method achieves a better LPIPS score than both BPG and VVC under an extremely low bitrate range ($<0.07$ bpp).
However, the increase of the bitrate brings constant quality improvement of conventional codecs BPG and VVC, while no significant improvement for the proposed method.
It validates that the proposed method aims to achieve better reconstruction quality at \textbf{extremely low bitrate} ($<0.1$ bpp) than conventional codecs.
}

\begin{figure}[!h]
    \centering
    \includegraphics[width=0.75\linewidth]{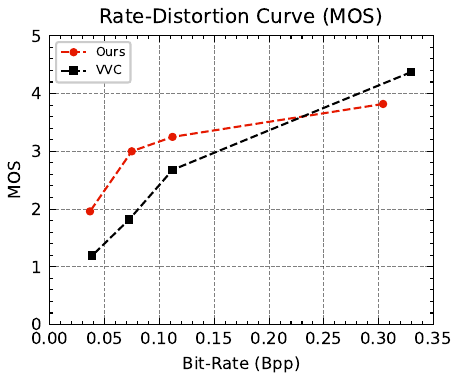}
    \caption{{\textbf{The mean opinion scores (MOS) of the proposed method and VVC at multi-level bitrate.} Higher score demonstrates better subjective reconstruction quality.}}
    \vspace{-4mm}
    \label{fig:mos}
\end{figure}

We also conduct subjective testing for evaluating image quality at various bitrate.
The double stimulus image quality assessment method is adopted to collect the subjective image quality scores from $25$ participants.
The testing consists of four typical bitrates ($0.037$ bpp, $0.074$ bpp, $0.112$ bpp, $0.3$ bpp) considering the bitrate range of the proposed method is $(0, 0.3)$ bpp as shown in Fig.~\ref{fig:RD}, where each test contains nine images selected randomly from the testing set from the testing set of CelebA-HQ dataset. 
The test images are presented by pairs, containing one uncompressed reference image and one random decoded image over all testing images.
The test is conducted on the screen with resolution of $1920\times 1080$ or higher and a HiDPI/Retina mode is not enabled.
The recommended score consists of value $\{1,2,3,4,5\}$ corresponding to the subjective quality level of \textit{Bad, Poor, Fair, Good} and \textit{Excellent}.
The mean opinion scores (MOS) are calculated for assessing the image quality of each method at each bitrate as following:
\begin{equation}
    MOS_{i}=\frac{1}{MN}\sum_{n=1}^{N}\sum_{m=1}^{M} s_{imn},
\end{equation}
where $N$ is the number of valid sujects and $M$ is the number of testing images at the bitrate $i$.  
The experimental results are shown in the rate-distortion curve of Fig. \ref{fig:mos}.
It can be seen that the proposed method obtains higher MOS than VVC at low bitrate testing ($0.037$ bpp, $0.074$ bpp, $0.112$ bpp), but a slightly lower score at high bitrate ($0.3$ bpp).
The subjective results validate the proposed method can achieve better visual reconstruction quality at an extremely low bitrate than traditional coding standard, consistent with the quantitative results of Fig. \ref{fig:RD}.
\begin{figure}[t]
    \centering 
    \vspace{-2mm}
    \includegraphics[width=1.0\linewidth]{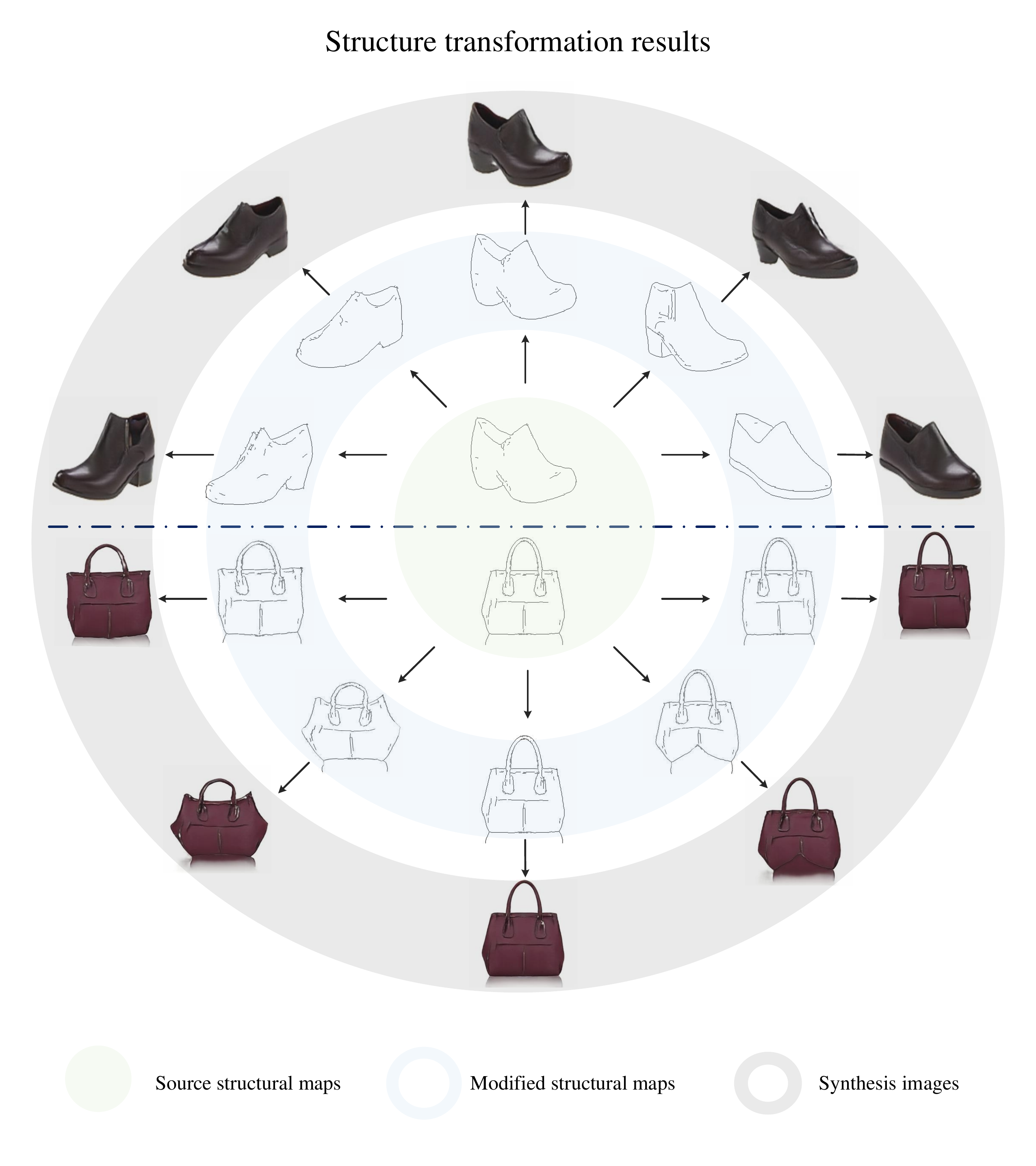}
    \caption{\textbf{Image synthesis results via modifying structural maps.} With the structural maps edited in optional ways, \textit{e.g.}, user interactive edition, image morphing algorithms, the texture layer can be flexibly rendered to fit the layout and shape of updated structural maps.}
    \label{fig:shape_transfer}
    \vspace{-4mm}
\end{figure}
\subsection{Application in Image Manipulation}
In our conceptual compression framework, images are explicitly disentangled to structure domain and texture domain, and reconstructed through integrating texture and structure with generator.
The structure and texture representations act as both raw frame data and inter-operable and manipulable visual features.
During the data-driven training, a continuous bidirectional mapping between deep latent representations and spatially-aware texture is well learned, allowing generator to synthesize corresponding content from any representation in texture latent space.
Benefited from the disentanglement of visual components and learned synthesis paradigm, the generator is able to progressively render the texture following the instruction of any given structural maps. 
Thus, besides high efficiency compression, the proposed framework can be also applied in image manipulation tasks via editing structure and texture representations. 
In particular, we can synthesize new texture in the way of changing the texture representations while constraining specific structural maps as image content.
On the other hand, we can also modify the structural maps while maintaining the image texture to satisfy particular needs for image manipulation.
The powerful capability of manipulating images flexibly in compression domain demonstrates the versatility of our conceptual compression framework in supporting both compression tasks and efficient content manipulation in the compressed domain compared to normal compression methods.
\subsubsection{Texture Synthesis}
The results of the texture synthesis results are shown in Fig.~\ref{fig:texture_transfer}. 
Attributed to analyzing and compressing images into compact disentangled structure and texture representations, the proposed method is able to process images in two independent layer before decoding.
Due to the unique association between specific representation and global texture of image instance, the implication of texture representation can be confirmed and reflected in the source images and synthesized images.
Thus, with all accessible images serving as a texture source library, the proposed method can synthesize different texture by replacing the texture representation with ones encoded from images of desired texture styles, and construct creative and attractive images.
The texture synthesis results in Fig.~\ref{fig:texture_transfer} show that our model has strong capacity of effectively capturing image global texture distribution, integrating representation layers without relying on encoder, and synthesizing pleasing target images, which suggest a great potential of proposed method in support of joint image compression and vision tasks. 
 
\begin{figure}[t]
    \centering
    \vspace{-4mm}
    \includegraphics[width=0.9\linewidth]{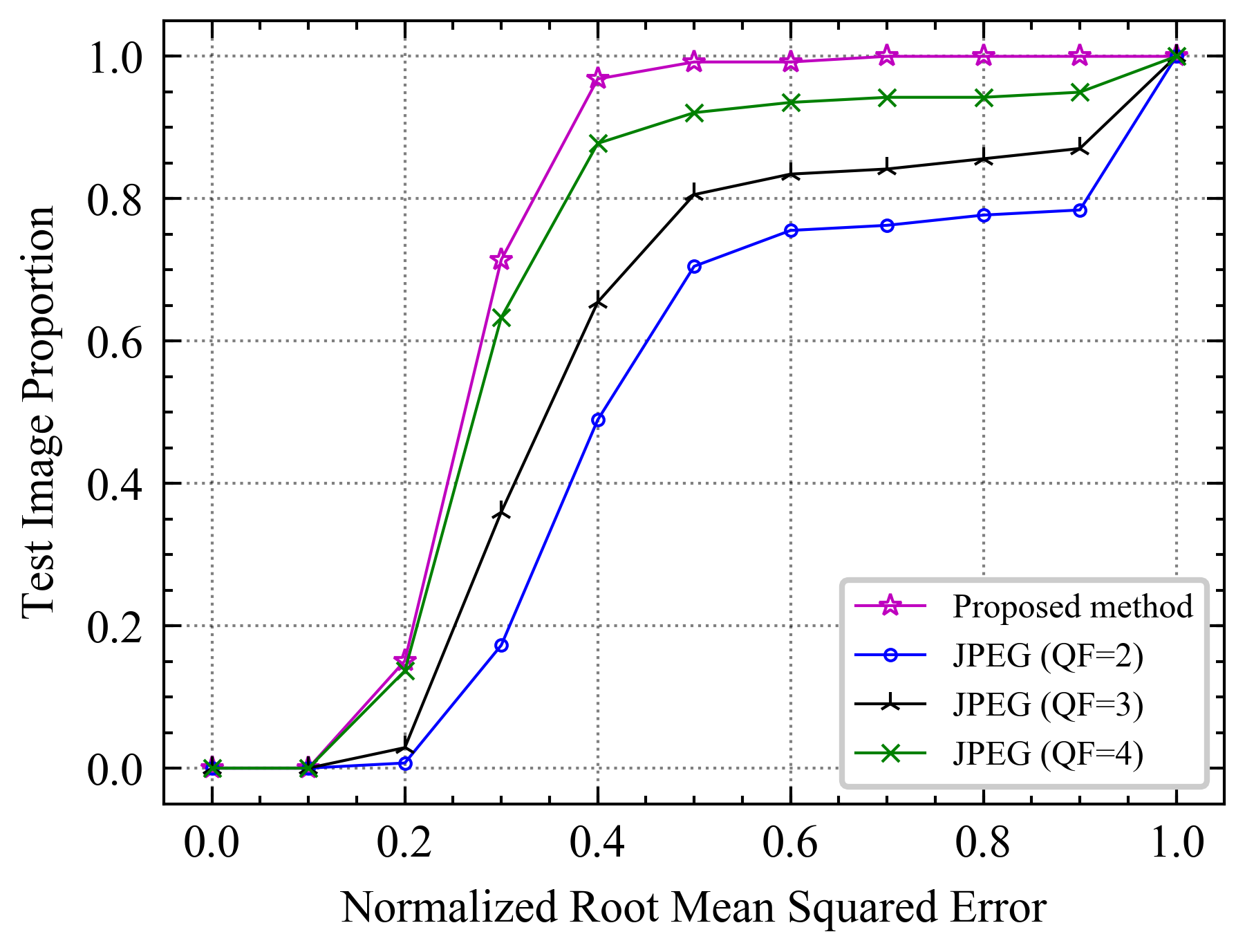}
    \caption{\textbf{Cumulative error distribution of facial landmark detection with JPEG under quality factor ${2,3,4}$ and proposed method.} It can be seen from the figure that the decoded test images from the proposed method ($97\%$), JPEG with quality factor 4 ($88\%$), JPEG with quality factor 3 ($65\%$), JPEG with quality factor 2 ($49\%$) achieve errors of less than 0.4 respectively.}
    \label{fig:RMSE}
    \vspace{-4mm}
\end{figure}
\subsubsection{Structure Modification}
We also demonstrate example results of applying the proposed method to structure modification in Fig.~\ref{fig:shape_transfer}.
Under the layered framework, the shape of images can be modified through editing the structural maps in optional ways, \textit{e.g.}, user interactive edition, image morphing algorithms.
The texture layer can be flexibly rendered fitting the layout and shape of updated structural maps and the results show that original texture can perfectly fit new shapes in synthesized images.
Structure modification is very useful in visual communication and image-based retrieval applications.
Moreover, operating machine vision tasks in the compressed domain is shown to be more efficient than operating it after decoding under low bit-rate scenarios~\cite{ma2018joint}.
Our experimental results also prove the potential advantage of the proposed method in performing high efficiency coding and machine vision tasks jointly.
\subsection{Advantage in Image Analysis}
To further verify the advantage of our framework in image analysis task, we also perform the facial landmark detection task on the decoded images.
Facial landmark detection~\cite{kazemi2014one} is carried out on the original images in CelebA-HQ testset and the detection results are served as ground truth.
For comparison, we perform landmark detection on the decoded images from JPEG and the proposed method respectively.
Then the normalized root mean squared error (NRMSE) between the detection results on compressed images and original images is calculated as quantitative metric.
Fig.~\ref{fig:RMSE} illustrates the cumulative error distribution of our method and JPEG with quality factor $\left \{ 2,3,4 \right \}$, where about $97\%$ of the test images reconstructed by proposed method have minor errors less than $0.4$.
It should be noted that the average bit-rate of our method and JPEG under $QF=1$ are $0.099$ bpp and $0.237$ bpp respectively.
Our method can achieve $58.1\%$ bits saving and $56.5\%$ accuracy improving of landmark detection task compared to JPEG under $QF=1$.
Above all, the face landmark detection results demonstrate robustness and accuracy of the proposed image compression method in image analysis tasks. 

In essence, in the proposed implementation scheme, the structural maps already contain key facial edges which act like dense landmarks, allowing straightforward content analysis without decoding.
Moreover, the concrete form of the structure representation can be further adjusted towards specific vision applications, leading to a unified framework for connecting machine vision and human perception through inter-operable conceptual representations.

\section{Discussion}
\subsection{Analysis of Incremental Reconstruction}
\begin{figure}[h]
    \centering
    \includegraphics[width=0.9\linewidth]{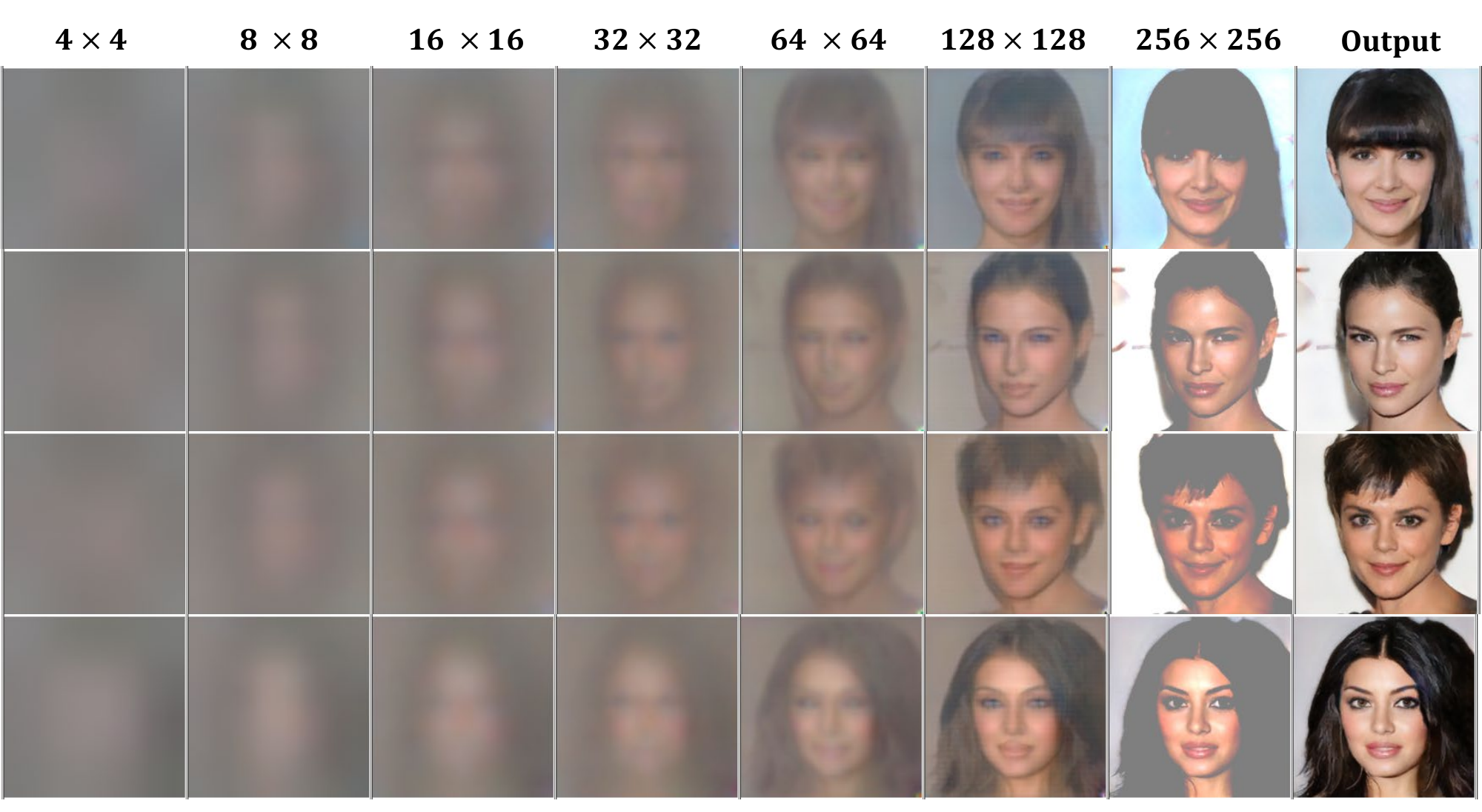}
    \caption{{\textbf{Visualization of the generative mechanism of HF-GAN.} We add the RGB feature (from toRGB) of the current Resblock to the upsampled RGB feature of the last Resblock as the incremental learning output for the current stage. The output of each stage from $4\times 4$ to $256\times 256$ in the hierarchical fusion generator network is visualized for illustrating the step-by-step synthesis process from rough structure to finer details.}}
    \label{fig:visualizaiton}
    \vspace{-4mm}
\end{figure}

The design of incremental reconstruction is motivated by previous works~\cite{karras2018progressive,karras2019style} with the common observation that high-resolution images is easier to learn in the coarse-to-fine manner in GANs. 
Fig.~\ref{fig:visualizaiton} illustrates the visualization of the output feature at each resolution stage in the proposed HF-GAN.
The step-by-step synthesis process focuses on rough architecture and boundaries first, and replenish finer details incrementally when the resolution raises up, validating that the incremental reconstruction strategy facilitates better quality of high-resolution image generation.

\subsection{Visual Quality and Signal Fidelity}
The difference and relationship between visual quality and signal fidelity have been studied theoretically in various research works~\cite{blau2018perception,liu2019classification}.
Signal fidelity is defined to evaluate how similar is the restored signal to the original signal~\cite{liu2019classification}. 
The representative assessment metrics include well-known mean-squared error (MSE) and its counterpart peak signal-to-noise ratio (PSNR), structural similarity (SSIM), and so on.
The quality of the image, which could be defined from a natural scene statistics perspective, is rooted in the widely accepted view that the vision system has adapted and evolved through the perception of the natural environment. As such, it has been empirically shown that the human visual system prefers natural images~\cite{blau2018perception}, and the 
perceptual quality refers to the extent that a reconstructed image looks like a valid natural image~\cite{liu2019classification,blau2018perception} rather than the signal similarity to the input.
The traditional codecs such as BPG mainly focus on signal fidelity.
By contrast, GANs employ the discriminator to implicitly characterize the natural scene statistics, such that the reconstructed images of the proposed conceptual compression scheme are closer to the natural image distribution.
This could inevitably damage the signal fidelity, which has been evidenced by a series of image restoration works~\cite{wang2021towards,yang2020hifacegan}.
The nature of the proposed method also limits the application scopes.
However, with the accelerated proliferation of digital devices, the vision-centered services that require high quality rending have been growing exponentially.
Moreover, in our future work, an engineering solution can be provided by adaptively adding a residual enhancement layer to satisfy the fidelity demand in practical application. 

\subsection{Generalization}
\begin{figure}[!h]
    \centering
    \vspace{-3mm}
    \subfloat[Decoded results on FFHQ dataset using the model trained on CelebA-HQ dataset]{
    \includegraphics[width=0.95\linewidth]{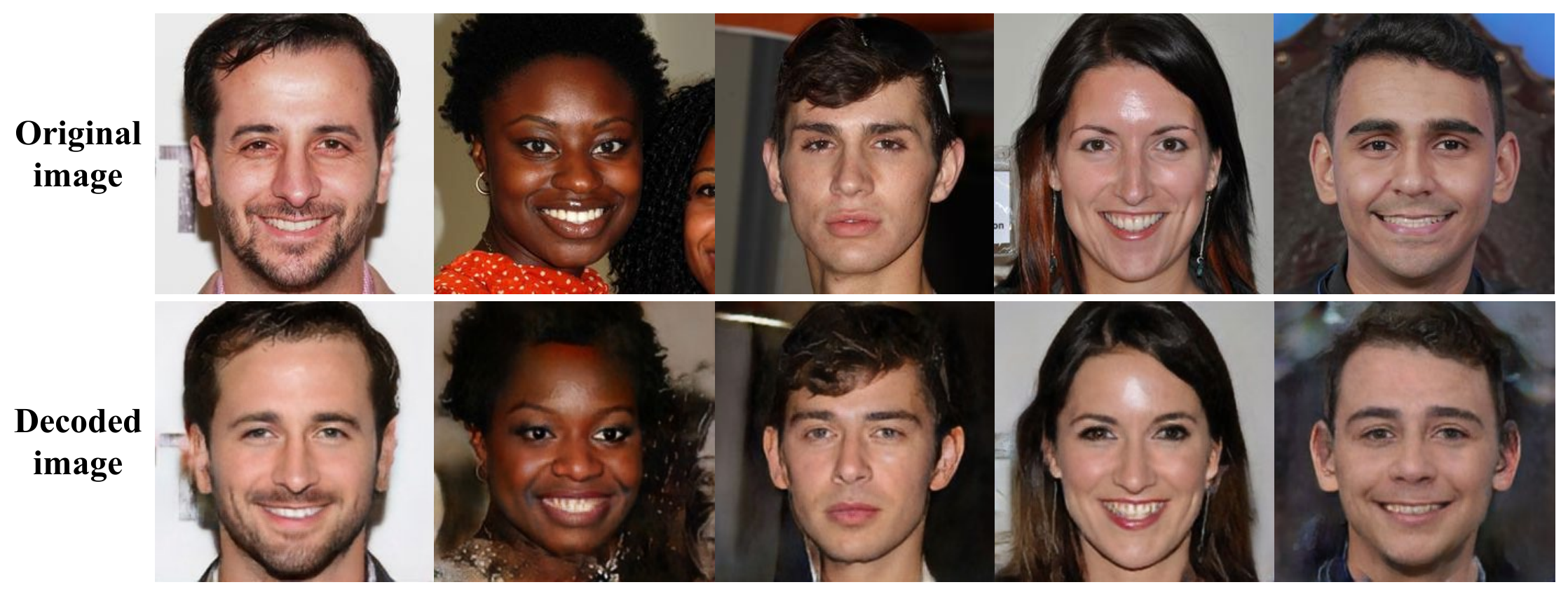}
    }
    
    \subfloat[Decoded results on ADE20K dataset using the model trained on the collected multiple seasons dataset]{
    \includegraphics[width=0.95\linewidth]{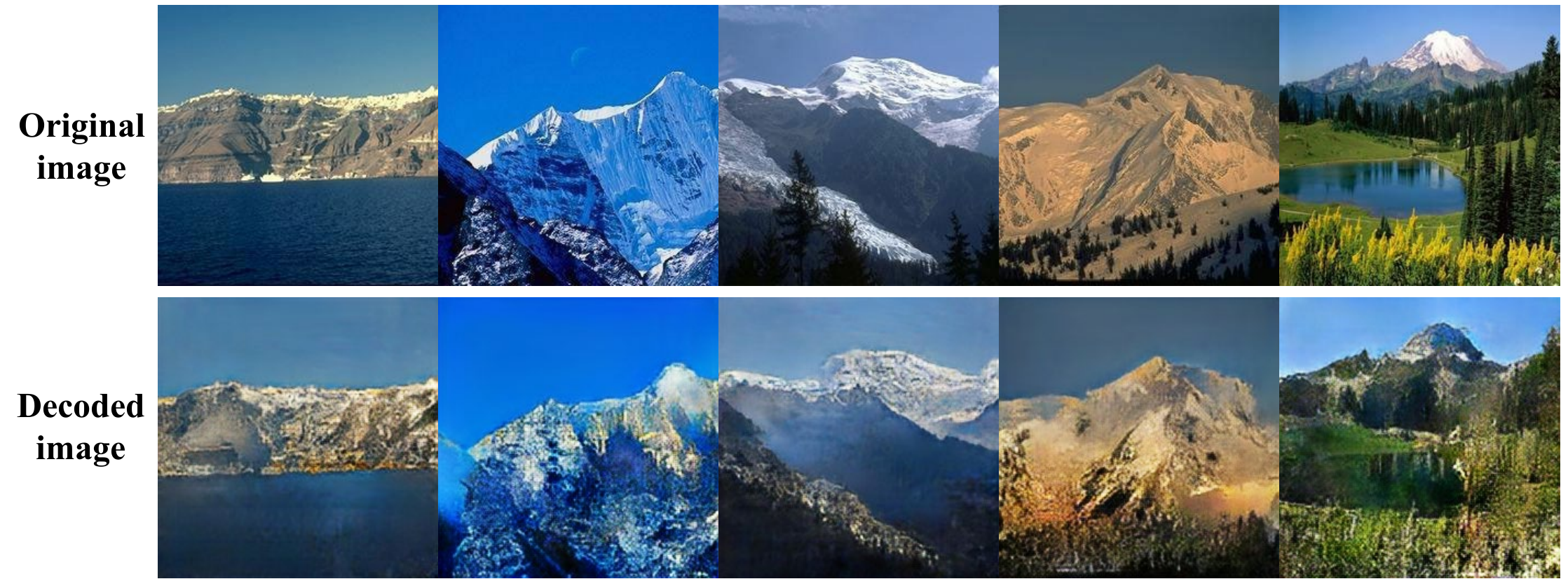}
    }
    \caption{\textbf{Generalization results on datasets with similar semantic objects.} The cross dataset application results demonstrate the generalization of our model on similar semantic object.}
    \label{fig:similar-semantic}
\end{figure}

\begin{figure}[!h]
    \centering
    \includegraphics[width=0.95\linewidth]{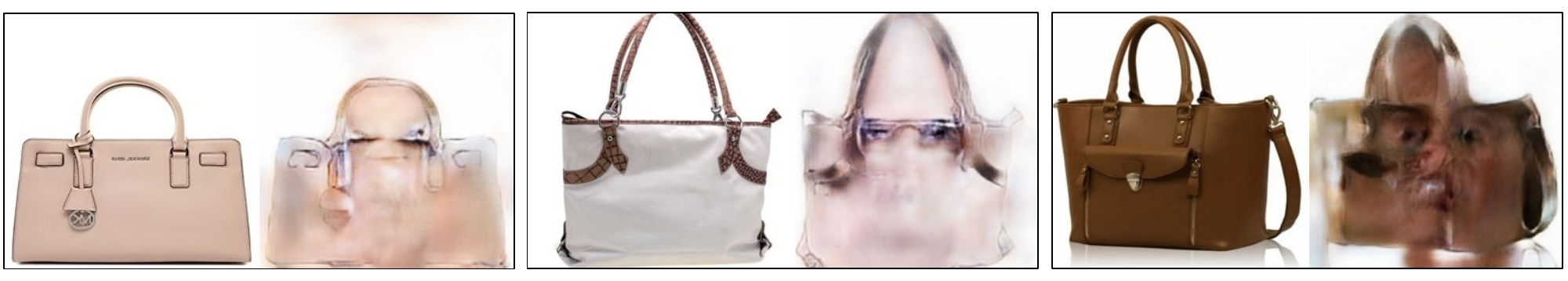}
    \caption{\textbf{Generalization results on datasets with large semantic gap.} We show the results of edge2handbags datasets on pretrained model using CelebA-HQ dataset, where the target texture can not be synthesized well with some semantic artifacts originated from the trained datasets, \eg facial semantic artifacts such as eyes when generating handbag textures.}
    \label{fig:cross-model-test}
    \vspace{-4mm}
\end{figure}

Different from the traditional engineering codecs based on linear transform and recently end-to-end learned codecs based on nonlinear transform~\cite{balle2020nonlinear}, the proposed framework lies in the deep generative models and learns a texture representation in a data-driven manner to capture the texture distribution of the training data domain. 
Thus, the texture representation can be generalized on images with similar distribution to the training domain.
Fig.~\ref{fig:similar-semantic} shows the experimental results on testing images from the FFHQ dataset and ADE20K dataset with models trained on CelebA-HQ and collected multiple seasons training set respectively, demonstrating that the model can generalize to the dataset with similar semantic object.
However, when applying to the dataset with a large semantic gap as shown in Fig.~\ref{fig:cross-model-test},
the model could not generate the texture as expected, and artifacts originating from the trained datasets (\eg eyes alike facial semantic artifacts), are perceived when generating handbag images.

While this approach has this merit, it is of considerable interest to improve the generalization capability of the proposed approach. There remains much work to be done in this direction. On one hand, we can improve the generalization capability based on domain generalization algorithms, such that data from different target domains can be efficiently compressed. Alternatively, we could also ensemble models from different domains as a more generalized codec. The images to be compressed are firstly classified into the specific domain, and the corresponding model is subsequently chosen for compression. The bitstream is composed of the representation of the images as well as the codec index, such that the images could be effectively decoded. It is worth mentioning that such an ensemble manner could well support myriad applications such as online shopping and video surveillance. It is our great desire to see that these emerging techniques could be used, especially in applications where the image content is loosely or strictly constrained.

\section{Conclusions and Future Works}
\label{conclusion}
This paper proposes a novel conceptual compression framework for efficient, interpretable and versatile visual data representation, leading to high efficiency compression, better visual reconstruction quality, increased flexibility in content manipulation and potential support for various vision tasks.
In particular, the proposed framework decomposes the image contents into structural and textural layers, and performs compression in two layers respectively.
To reconstruct the original image from the compressed layered features, a hierarchical fusion GAN is proposed to integrate texture layer and structure layer in a disentangled fashion.
Qualitative and quantitative results show that the proposed method can reconstruct high visual quality images with better structural fidelity and aesthetic sensibility.
The advantages of our conceptual compression framework are also verified in compression, content manipulation and  analysis tasks through extensive experiments.

Despite the current achievements, the conceptual compression will further benefit from the enhanced rate-distortion optimization schemes and better layered decomposition of image contents. Furthermore, extending the proposed conceptual image compression framework to video domain is another promising direction, where effective spatio-temporal representation learning plays a crucial part in a wide range of video analysis tasks, such as action recognition, tracking, summarization, and editing.

\section*{Acknowledgment}
The authors would like to thank the associate editor and anonymous reviewers for their valuable comments that significantly helped them in improving the quality of the paper. Assistance for the development of this research work provided by Prof. Shanshe Wang, Prof. Xinfeng Zhang and Dr. Jiguo Li was greatly appreciated.

\ifCLASSOPTIONcaptionsoff
  \newpage
\fi
\bibliographystyle{IEEEbib}
\bibliography{refs}

\begin{thebibliography}{10}

\bibitem{kruger2012deep}
Norbert Kruger, Peter Janssen, Sinan Kalkan, Markus Lappe, Ales Leonardis,
  Justus Piater, Antonio~J Rodriguez-Sanchez, and Laurenz Wiskott,
\newblock ``Deep hierarchies in the primate visual cortex: What can we learn
  for computer vision?,''
\newblock {\em IEEE Transactions on Pattern Analysis and Machine Intelligence},
  vol. 35, no. 8, pp. 1847--1871, 2012.

\bibitem{zhang2020connecting}
Yizhen Zhang, Kuan Han, Robert Worth, and Zhongming Liu,
\newblock ``Connecting concepts in the brain by mapping cortical
  representations of semantic relations,''
\newblock {\em Nature Communications}, vol. 11, no. 1, pp. 1--13, 2020.

\bibitem{pennebaker1992jpeg}
William~B Pennebaker and Joan~L Mitchell,
\newblock {\em {JPEG}: Still image data compression standard},
\newblock Springer Science \& Business Media, 1992.

\bibitem{sullivan2012overview}
Gary~J Sullivan, Jens-Rainer Ohm, Woo-Jin Han, and Thomas Wiegand,
\newblock ``Overview of the high efficiency video coding ({HEVC}) standard,''
\newblock {\em IEEE Transactions on Circuits and Systems for Video Technology},
  vol. 22, no. 12, pp. 1649--1668, 2012.

\bibitem{balle2019end}
Johannes Ball{\'e}, Valero Laparra, and Eero Simoncelli,
\newblock ``End-to-end optimized image compression,''
\newblock in {\em 5th International Conference on Learning Representations
  ({ICLR})}, 2017.

\bibitem{minnen2018joint}
David Minnen, Johannes Ball{\'e}, and George~D Toderici,
\newblock ``Joint autoregressive and hierarchical priors for learned image
  compression,''
\newblock in {\em Advances in Neural Information Processing Systems ({NIPS})},
  2018, pp. 10771--10780.

\bibitem{marr1982computational}
David Marr,
\newblock ``Vision: A computational investigation into the human representation
  and processing of visual information,''
\newblock {\em WH San Francisco: Freeman and Company}, vol. 1, no. 2, 1982.

\bibitem{guo2007primal}
Cheng-en Guo, Song-Chun Zhu, and Ying~Nian Wu,
\newblock ``Primal sketch: Integrating structure and texture,''
\newblock {\em Computer Vision and Image Understanding}, vol. 106, no. 1, pp.
  5--19, 2007.

\bibitem{kingma2014auto}
Diederik~P Kingma and Max Welling,
\newblock ``Auto-encoding variational bayes,''
\newblock {\em stat}, vol. 1050, pp. 1, 2014.

\bibitem{goodfellow2014generative}
Ian Goodfellow, Jean Pouget-Abadie, Mehdi Mirza, Bing Xu, David Warde-Farley,
  Sherjil Ozair, Aaron Courville, and Yoshua Bengio,
\newblock ``Generative adversarial nets,''
\newblock in {\em Advances in Neural Information Processing Systems ({NIPS})},
  2014, pp. 2672--2680.

\bibitem{gregor2016towards}
Karol Gregor, Frederic Besse, Danilo~Jimenez Rezende, Ivo Danihelka, and Daan
  Wierstra,
\newblock ``Towards conceptual compression,''
\newblock in {\em Advances in Neural Information Processing Systems ({NIPS})},
  2016, pp. 3549--3557.

\bibitem{santurkar2018generative}
Shibani Santurkar, David Budden, and Nir Shavit,
\newblock ``Generative compression,''
\newblock in {\em Picture Coding Symposium (PCS)}. IEEE, 2018, pp. 258--262.

\bibitem{gao2021digital}
Wen Gao, Siwei Ma, Lingyu Duan, Yonghong Tian, Peiyin Xing, Yaowei Wang,
  Shanshe Wang, Huizhu Jia, and Tiejun Huang,
\newblock ``Digital retina: A way to make the city brain more efficient by
  visual coding,''
\newblock {\em IEEE Transactions on Circuits and Systems for Video Technology},
  vol. 31, no. 11, pp. 4147--4161, 2021.

\bibitem{wang2018high}
Ting-Chun Wang, Ming-Yu Liu, Jun-Yan Zhu, Andrew Tao, Jan Kautz, and Bryan
  Catanzaro,
\newblock ``High-resolution image synthesis and semantic manipulation with
  conditional {GANs},''
\newblock in {\em Proceedings of the IEEE Conference on Computer Vision and
  Pattern Recognition ({CVPR})}, 2018, pp. 8798--8807.

\bibitem{akbari2019dsslic}
Mohammad Akbari, Jie Liang, and Jingning Han,
\newblock ``{DSSLIC}: Deep semantic segmentation-based layered image
  compression,''
\newblock in {\em IEEE International Conference on Acoustics, Speech and Signal
  Processing (ICASSP)}, 2019, pp. 2042--2046.

\bibitem{ma2018joint}
Siwei Ma, Xiang Zhang, Shiqi Wang, Xinfeng Zhang, Chuanmin Jia, and Shanshe
  Wang,
\newblock ``Joint feature and texture coding: Toward smart video representation
  via front-end intelligence,''
\newblock {\em IEEE Transactions on Circuits and Systems for Video Technology},
  vol. 29, no. 10, pp. 3095--3105, 2018.

\bibitem{yu2014fine}
Aron Yu and Kristen Grauman,
\newblock ``Fine-grained visual comparisons with local learning,''
\newblock in {\em Proceedings of the IEEE Conference on Computer Vision and
  Pattern Recognition ({CVPR})}, 2014, pp. 192--199.

\bibitem{zhu2016generative}
Jun-Yan Zhu, Philipp Kr{\"a}henb{\"u}hl, Eli Shechtman, and Alexei~A Efros,
\newblock ``Generative visual manipulation on the natural image manifold,''
\newblock in {\em European Conference on Computer Vision ({ECCV})}. Springer,
  2016, pp. 597--613.

\bibitem{karras2018progressive}
Tero Karras, Timo Aila, Samuli Laine, and Jaakko Lehtinen,
\newblock ``Progressive growing of gans for improved quality, stability, and
  variation,''
\newblock in {\em International Conference on Learning Representations
  ({ICLR})}, 2018.

\bibitem{rabbani2002jpeg2000}
Majid Rabbani,
\newblock ``{JPEG2000}: Image compression fundamentals, standards and
  practice,''
\newblock {\em Journal of Electronic Imaging}, vol. 11, no. 2, pp. 286, 2002.

\bibitem{bellard2015bpg}
Fabrice Bellard,
\newblock ``Bpg image format,''
\newblock {\em URL https://bellard.org/bpg}, 2015.

\bibitem{wiegand2003overview}
Thomas Wiegand, Gary~J Sullivan, Gisle Bjontegaard, and Ajay Luthra,
\newblock ``Overview of the {H.264/AVC} video coding standard,''
\newblock {\em IEEE Transactions on Circuits and Systems for Video Technology},
  vol. 13, no. 7, pp. 560--576, 2003.

\bibitem{gao2014overview}
Wen Gao and Siwei Ma,
\newblock ``An overview of {AVS2} standard,''
\newblock in {\em Advanced Video Coding Systems}, pp. 35--49. Springer, 2014.

\bibitem{AVS3}
Jiaqi Zhang, Chuanmin Jia, Meng Lei, Shanshe Wang, Siwei Ma, and Wen Gao,
\newblock ``Recent development of avs video coding standard: {AVS3},''
\newblock in {\em 2019 Picture Coding Symposium (PCS)}, 2019, pp. 1--5.

\bibitem{ma2019image}
Siwei Ma, Xinfeng Zhang, Chuanmin Jia, Zhenghui Zhao, Shiqi Wang, and Shanshe
  Wang,
\newblock ``Image and video compression with neural networks: A review,''
\newblock {\em IEEE Transactions on Circuits and Systems for Video Technology},
  vol. 30, no. 6, pp. 1683--1698, 2019.

\bibitem{salimans2017pixelcnn++}
Tim Salimans, Andrej Karpathy, Xi~Chen, and Diederik~P Kingma,
\newblock ``{PixelCNN++}: Improving the {pixelCNN} with discretized logistic
  mixture likelihood and other modifications,''
\newblock {\em International Conference on Learning Representations ({ICLR})},
  2017.

\bibitem{liu2020deep}
Dong Liu, Yue Li, Jianping Lin, Houqiang Li, and Feng Wu,
\newblock ``Deep learning-based video coding: a review and a case study,''
\newblock {\em ACM Computing Surveys (CSUR)}, vol. 53, no. 1, pp. 1--35, 2020.

\bibitem{toderici2015variable}
George Toderici, Sean~M O'Malley, Sung~Jin Hwang, Damien Vincent, David Minnen,
  Shumeet Baluja, Michele Covell, and Rahul Sukthankar,
\newblock ``Variable rate image compression with recurrent neural networks,''
\newblock {\em International Conference on Learning Representations ({ICLR})},
  2016.

\bibitem{agustsson2019generative}
Eirikur Agustsson, Michael Tschannen, Fabian Mentzer, Radu Timofte, and Luc~Van
  Gool,
\newblock ``Generative adversarial networks for extreme learned image
  compression,''
\newblock in {\em Proceedings of the IEEE International Conference on Computer
  Vision ({ICCV})}, 2019, pp. 221--231.

\bibitem{zhang2017deep}
Qingyu Zhang, Dong Liu, and Houqiang Li,
\newblock ``Deep network-based image coding for simultaneous compression and
  retrieval,''
\newblock in {\em IEEE International Conference on Image Processing (ICIP)},
  2017, pp. 405--409.

\bibitem{luo2018deepsic}
Sihui Luo, Yezhou Yang, Yanling Yin, Chengchao Shen, Ya~Zhao, and Mingli Song,
\newblock ``{DeepSIC}: Deep semantic image compression,''
\newblock in {\em International Conference on Neural Information Processing}.
  Springer, 2018, pp. 96--106.

\bibitem{torfason2018towards}
Robert Torfason, Fabian Mentzer, Eirikur Agustsson, Michael Tschannen, Radu
  Timofte, and Luc Van~Gool,
\newblock ``Towards image understanding from deep compression without
  decoding,''
\newblock in {\em International Conference on Learning Representations
  ({ICLR})}, 2018.

\bibitem{chang2019layered}
Jianhui Chang, Qi~Mao, Zhenghui Zhao, Shanshe Wang, Shiqi Wang, Hong Zhu, and
  Siwei Ma,
\newblock ``Layered conceptual image compression via deep semantic synthesis,''
\newblock in {\em IEEE International Conference on Image Processing (ICIP)},
  2019, pp. 694--698.

\bibitem{chang2021thousand}
Jianhui Chang, Zhenghui Zhao, Lingbo Yang, Chuanmin Jia, Jian Zhang, and Siwei
  Ma,
\newblock ``Thousand to one: Semantic prior modeling for conceptual coding,''
\newblock in {\em 2021 IEEE International Conference on Multimedia and Expo
  (ICME)}. IEEE, 2021, pp. 1--6.

\bibitem{hu2020towards}
Yueyu Hu, Shuai Yang, Wenhan Yang, Ling-Yu Duan, and Jiaying Liu,
\newblock ``Towards coding for human and machine vision: A scalable image
  coding approach,''
\newblock in {\em IEEE International Conference on Multimedia and Expo (ICME)}.
  IEEE, 2020, pp. 1--6.

\bibitem{xie2015holistically}
Saining Xie and Zhuowen Tu,
\newblock ``Holistically-nested edge detection,''
\newblock in {\em Proceedings of the IEEE international Conference on Computer
  Vision ({ICCV})}, 2015, pp. 1395--1403.

\bibitem{ding2001canny}
Lijun Ding and Ardeshir Goshtasby,
\newblock ``On the canny edge detector,''
\newblock {\em Pattern Recognition}, vol. 34, no. 3, pp. 721--725, 2001.

\bibitem{6783726}
Weijia Zhu, Wenpeng Ding, Jizheng Xu, Yunhui Shi, and Baocai Yin,
\newblock ``Screen content coding based on {HEVC} framework,''
\newblock {\em IEEE Transactions on Multimedia}, vol. 16, no. 5, pp.
  1316--1326, Aug 2014.

\bibitem{haris2018deep}
Muhammad Haris, Gregory Shakhnarovich, and Norimichi Ukita,
\newblock ``Deep back-projection networks for super-resolution,''
\newblock in {\em Proceedings of the IEEE Conference on Computer Vision and
  Pattern Recognition ({CVPR})}, 2018, pp. 1664--1673.

\bibitem{he2016deep}
Kaiming He, Xiangyu Zhang, Shaoqing Ren, and Jian Sun,
\newblock ``Deep residual learning for image recognition,''
\newblock in {\em Proceedings of the IEEE Conference on Computer Vision and
  Pattern Recognition ({CVPR})}, 2016, pp. 770--778.

\bibitem{witten1987arithmetic}
Ian~H Witten, Radford~M Neal, and John~G Cleary,
\newblock ``Arithmetic coding for data compression,''
\newblock {\em Communications of the ACM}, vol. 30, no. 6, pp. 520--540, 1987.

\bibitem{karras2020analyzing}
Tero Karras, Samuli Laine, Miika Aittala, Janne Hellsten, Jaakko Lehtinen, and
  Timo Aila,
\newblock ``Analyzing and improving the image quality of stylegan,''
\newblock in {\em Proceedings of the IEEE/CVF Conference on Computer Vision and
  Pattern Recognition ({CVPR})}, 2020, pp. 8110--8119.

\bibitem{huang2017arbitrary}
Xun Huang and Serge Belongie,
\newblock ``Arbitrary style transfer in real-time with adaptive instance
  normalization,''
\newblock in {\em Proceedings of the IEEE International Conference on Computer
  Vision}, 2017, pp. 1501--1510.

\bibitem{wang2004image}
Zhou Wang, Alan~C Bovik, Hamid~R Sheikh, and Eero~P Simoncelli,
\newblock ``Image quality assessment: from error visibility to structural
  similarity,''
\newblock {\em IEEE Transactions on Image Processing}, vol. 13, no. 4, pp.
  600--612, 2004.

\bibitem{johnson2016perceptual}
Justin Johnson, Alexandre Alahi, and Li~Fei-Fei,
\newblock ``Perceptual losses for real-time style transfer and
  super-resolution,''
\newblock in {\em European Conference on Computer Vision (ECCV)}. Springer,
  2016, pp. 694--711.

\bibitem{simonyan2014very}
Karen Simonyan and Andrew Zisserman,
\newblock ``Very deep convolutional networks for large-scale image
  recognition,''
\newblock {\em International Conference on Learning Representations ({ICLR})},
  2015.

\bibitem{mao2017least}
Xudong Mao, Qing Li, Haoran Xie, Raymond~YK Lau, Zhen Wang, and Stephen
  Paul~Smolley,
\newblock ``Least squares generative adversarial networks,''
\newblock in {\em Proceedings of the IEEE International Conference on Computer
  Vision ({ICCV})}, 2017, pp. 2794--2802.

\bibitem{zhu2017unpaired}
Jun-Yan Zhu, Taesung Park, Phillip Isola, and Alexei~A Efros,
\newblock ``Unpaired image-to-image translation using cycle-consistent
  adversarial networks,''
\newblock in {\em Proceedings of the IEEE International Conference on Computer
  Vision ({ICCV})}, 2017, pp. 2223--2232.

\bibitem{anoosheh2018combogan}
Asha Anoosheh, Eirikur Agustsson, Radu Timofte, and Luc Van~Gool,
\newblock ``Combogan: Unrestrained scalability for image domain translation,''
\newblock in {\em Proceedings of the IEEE Conference on Computer Vision and
  Pattern Recognition Workshops}, 2018, pp. 783--790.

\bibitem{bross2021overview}
Benjamin Bross, Ye-Kui Wang, Yan Ye, Shan Liu, Jianle Chen, Gary~J Sullivan,
  and Jens-Rainer Ohm,
\newblock ``Overview of the versatile video coding ({VVC}) standard and its
  applications,''
\newblock {\em IEEE Transactions on Circuits and Systems for Video Technology},
  vol. 31, no. 10, pp. 3736--3764, 2021.

\bibitem{wang2011ssim}
Shiqi Wang, Abdul Rehman, Zhou Wang, Siwei Ma, and Wen Gao,
\newblock ``{SSIM}-motivated rate-distortion optimization for video coding,''
\newblock {\em IEEE Transactions on Circuits and Systems for Video Technology},
  vol. 22, no. 4, pp. 516--529, 2011.

\bibitem{helmrich2019perceptually}
Christian~R Helmrich, Sebastian Bosse, Mischa Siekmann, Heiko Schwarz, Detlev
  Marpe, and Thomas Wiegand,
\newblock ``Perceptually optimized bit-allocation and associated distortion
  measure for block-based image or video coding,''
\newblock in {\em Proceedings of the IEEE Conference on Data Compression
  Conference (DCC)}, 2019, pp. 172--181.

\bibitem{sun2019perceptual}
Xuebin Sun, Han Ma, Weixun Zuo, and Ming Liu,
\newblock ``Perceptual-based {HEVC} intra coding optimization using deep
  convolution networks,''
\newblock {\em IEEE Access}, vol. 7, pp. 56308--56316, 2019.

\bibitem{zhang2018unreasonable}
Richard Zhang, Phillip Isola, Alexei~A Efros, Eli Shechtman, and Oliver Wang,
\newblock ``The unreasonable effectiveness of deep features as a perceptual
  metric,''
\newblock in {\em Proceedings of the IEEE Conference on Computer Vision and
  Pattern Recognition ({CVPR})}, 2018, pp. 586--595.

\bibitem{ding2020image}
Keyan Ding, Kede Ma, Shiqi Wang, and Eero~P Simoncelli,
\newblock ``Image quality assessment: Unifying structure and texture
  similarity,''
\newblock {\em IEEE Transactions on Pattern Analysis \& Machine Intelligence},
  pp. 1--1, 2020.

\bibitem{li2021quality}
Yang Li, Shiqi Wang, Xinfeng Zhang, Shanshe Wang, Siwei Ma, and Yue Wang,
\newblock ``Quality assessment of end-to-end learned image compression: The
  benchmark and objective measure,''
\newblock in {\em Proceedings of the 29th ACM International Conference on
  Multimedia}, 2021, pp. 4297--4305.

\bibitem{kazemi2014one}
Vahid Kazemi and Josephine Sullivan,
\newblock ``One millisecond face alignment with an ensemble of regression
  trees,''
\newblock in {\em Proceedings of the IEEE Conference on Computer Vision and
  Pattern Recognition ({CVPR})}, 2014, pp. 1867--1874.

\bibitem{karras2019style}
Tero Karras, Samuli Laine, and Timo Aila,
\newblock ``A style-based generator architecture for generative adversarial
  networks,''
\newblock in {\em Proceedings of the IEEE Conference on Computer Vision and
  Pattern Recognition (CVPR)}, 2019, pp. 4401--4410.

\bibitem{blau2018perception}
Yochai Blau and Tomer Michaeli,
\newblock ``The perception-distortion tradeoff,''
\newblock in {\em Proceedings of the IEEE Conference on Computer Vision and
  Pattern Recognition (CVPR)}, 2018, pp. 6228--6237.

\bibitem{liu2019classification}
Dong Liu, Haochen Zhang, and Zhiwei Xiong,
\newblock ``On the classification-distortion-perception tradeoff,''
\newblock {\em Advances in Neural Information Processing Systems (NeurIPS)},
  vol. 32, pp. 1206--1215, 2019.

\bibitem{wang2021towards}
Xintao Wang, Yu~Li, Honglun Zhang, and Ying Shan,
\newblock ``Towards real-world blind face restoration with generative facial
  prior,''
\newblock in {\em Proceedings of the IEEE/CVF Conference on Computer Vision and
  Pattern Recognition}, 2021, pp. 9168--9178.

\bibitem{yang2020hifacegan}
Lingbo Yang, Shanshe Wang, Siwei Ma, Wen Gao, Chang Liu, Pan Wang, and Peiran
  Ren,
\newblock ``Hiface{GAN}: Face renovation via collaborative suppression and
  replenishment,''
\newblock in {\em Proceedings of the 28th ACM International Conference on
  Multimedia}, 2020, pp. 1551--1560.

\bibitem{balle2020nonlinear}
Johannes Ball{\'e}, Philip~A Chou, David Minnen, Saurabh Singh, Nick Johnston,
  Eirikur Agustsson, Sung~Jin Hwang, and George Toderici,
\newblock ``Nonlinear transform coding,''
\newblock {\em IEEE Journal of Selected Topics in Signal Processing}, vol. 15,
  no. 2, pp. 339--353, 2020.

\end{thebibliography}

\begin{IEEEbiography}[{\includegraphics[width=1in,height=1.25in,clip,keepaspectratio]{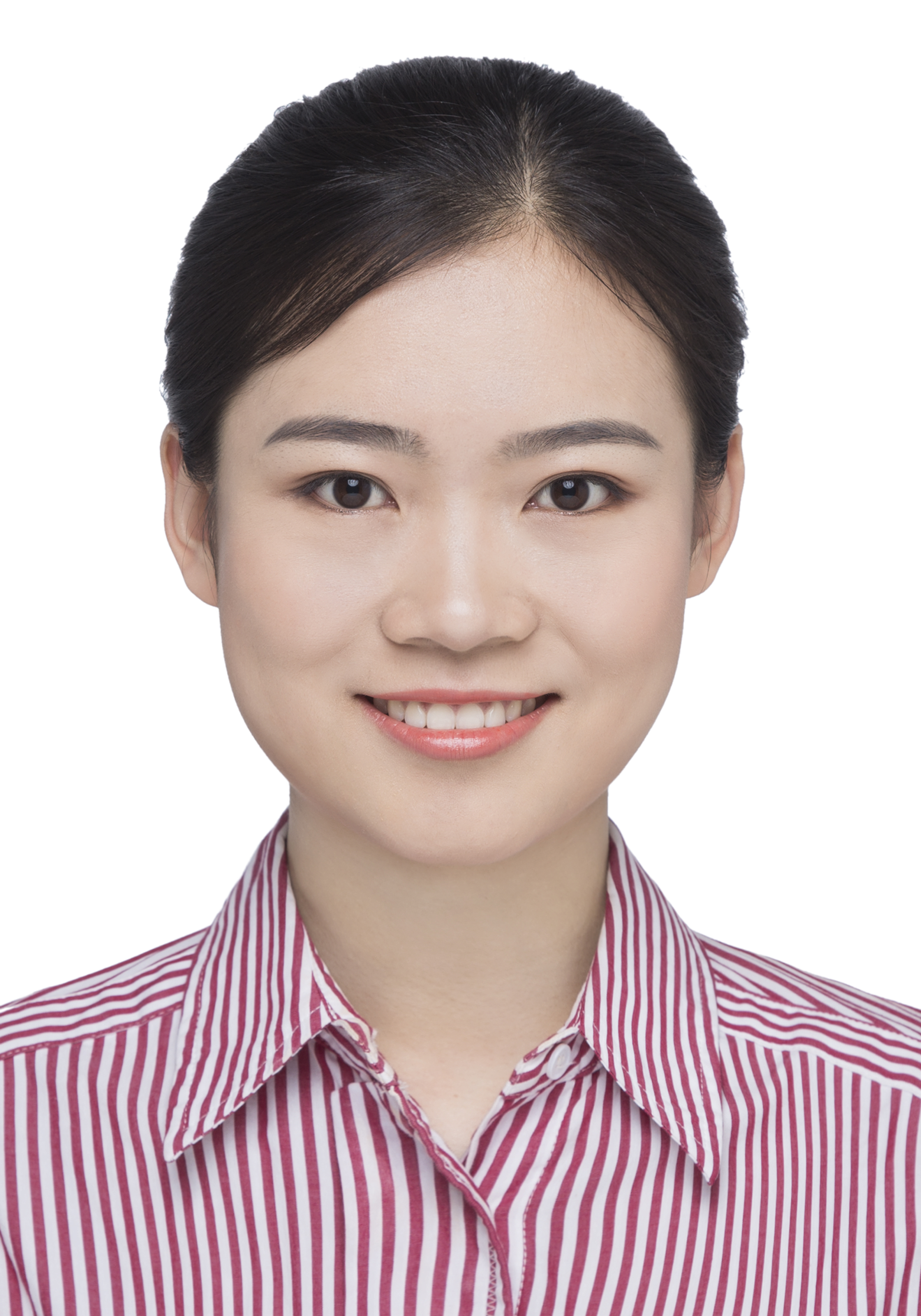}}]{Jianhui Chang} received B.E. degree in computer science and technology from China University of Mining and Technology (Beijing), China, in 2019. She is currently pursuing the Ph.D. degree with Institute of Digital Media in Peking University, Beijing, China. She has been dedicating herself to the fundamental theories and frameworks for visual data compression. Her research interests include  deep generative models, image synthesis, image/video compression, and video coding for machine. 
\end{IEEEbiography}

 \begin{IEEEbiography}[{\includegraphics[width=1in,height=1.25in,clip,keepaspectratio]{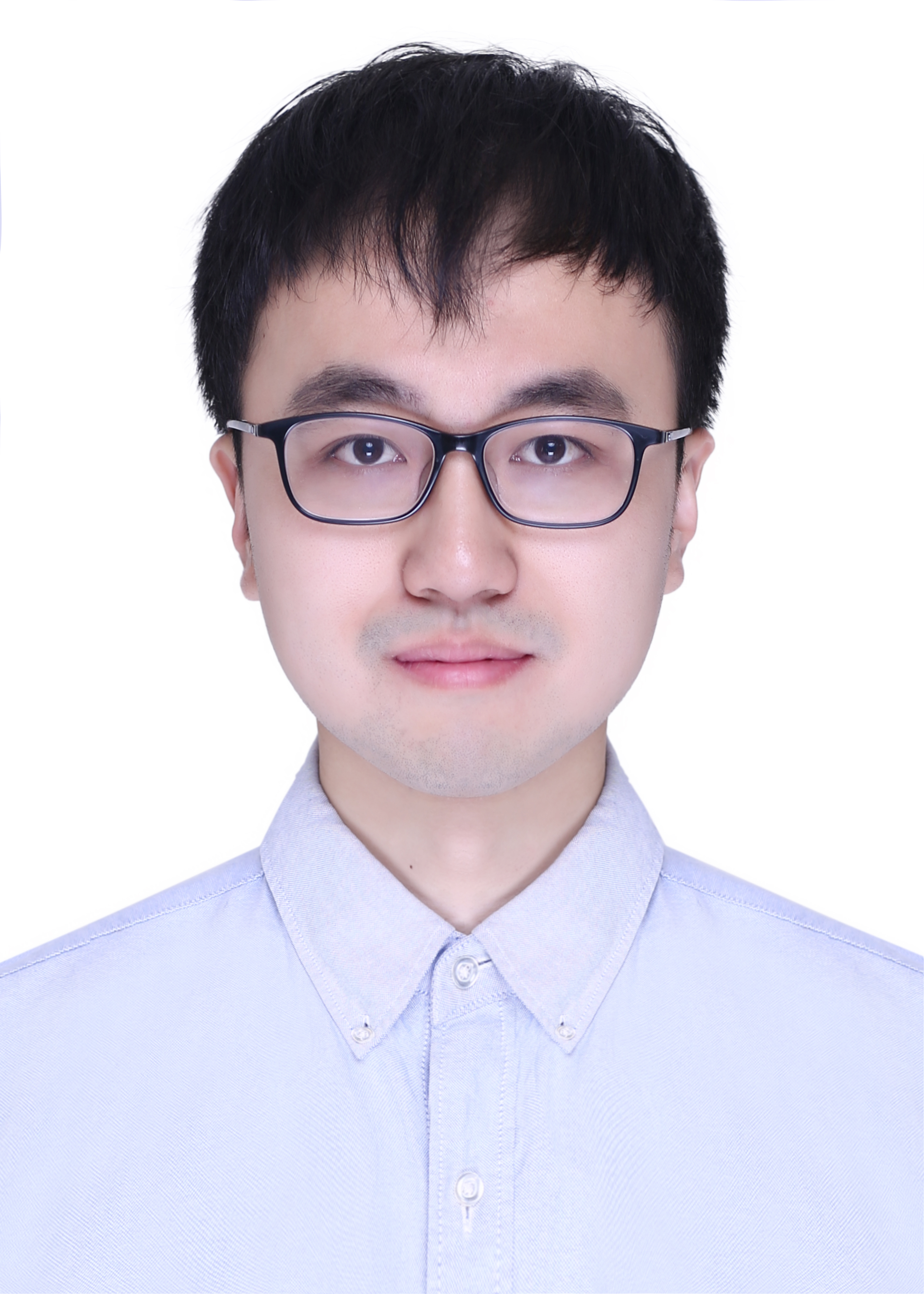}}]
{Zhenghui Zhao} received the B.S. degree in the School of Mathematical Sciences from Dalian University of Technology in 2015. He received the Ph.D. degree in the School of Mathematical Sciences from Peking University in 2021. Currently he is working as an algorithm engineer in the Alibaba Cloud Computing Co. Ltd. His research interests include video coding standards, hardware-oriented algorithms of video encoders and deep learning based compression algorithms.
\end{IEEEbiography}

\begin{IEEEbiography}[{\includegraphics[width=1in,height=1.25in,clip,keepaspectratio]{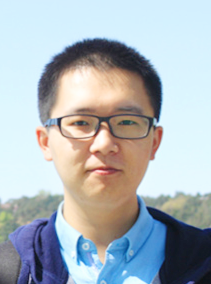}}]{Chuanmin Jia}
   received B.E. degree in computer science from Beijing University of Posts and Telecommunications, Beijing, China, in 2015 and the Ph.D. degree in computer application technology from Peking University, Beijing, China, in 2020. He was a visiting student with Video Lab, New York University, NY, USA, in 2018. He is currently working as Boya postdoc fellow with the Department of Computer Science, Peking University, Beijing, China. His research interests include image/video compression, multimedia signal processing and analysis. He is the recipient of Best Paper Award of Pacific-Rim Conference on Multimedia in 2017, Best Student Paper Award of IEEE International Conference on Multimedia Information Processing and Retrieval 2019, and also the co-recipient of Best Paper Award of IEEE Multimedia Magazine in 2018.
  \end{IEEEbiography}
 
\begin{IEEEbiography}[{\includegraphics[width=1in,height=1.25in,clip,keepaspectratio]{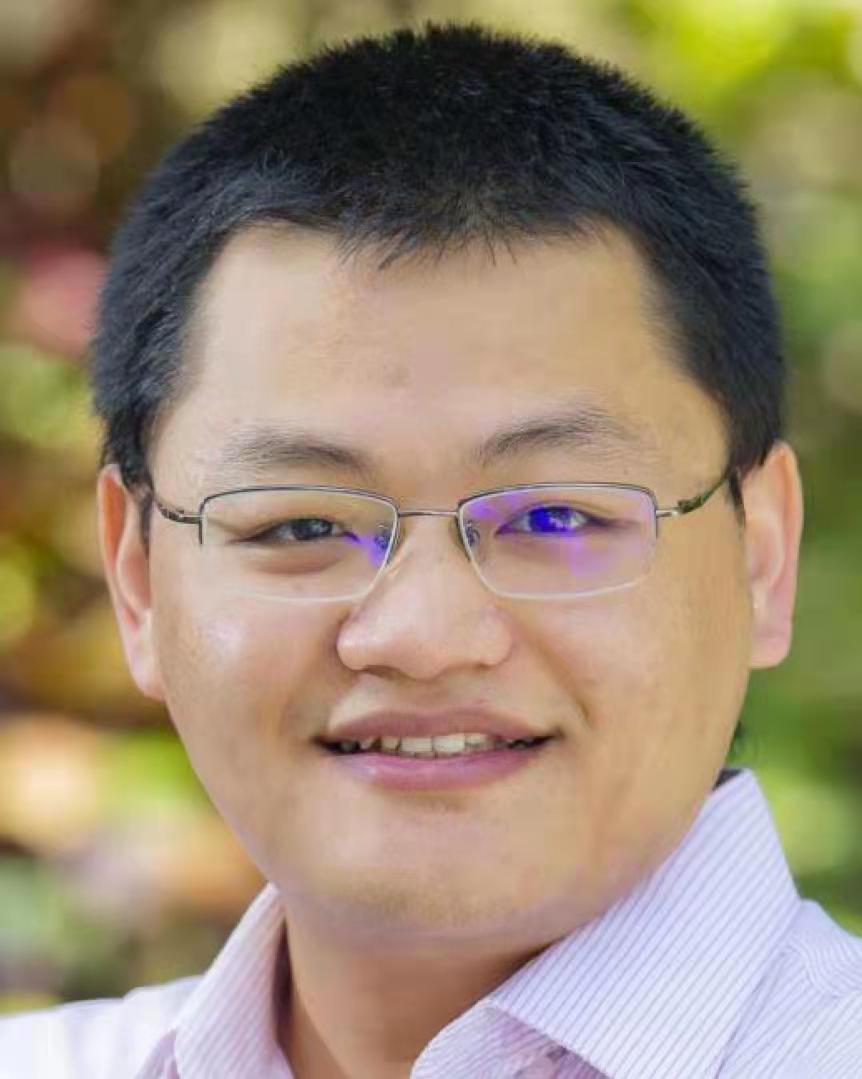}}]
{Shiqi Wang} (Senior Member, IEEE) received the B.S. degree in computer science from the Harbin Institute of Technology in 2008 and the Ph.D. degree in computer application technology from Peking University in 2014. From 2014 to 2016, he was a Post-Doctoral Fellow with the Department of Electrical and Computer Engineering, University of Waterloo, Waterloo, ON, Canada. From 2016 to 2017, he was a Research Fellow with the Rapid-Rich Object Search Laboratory, Nanyang Technological University, Singapore. He is currently an Assistant Professor with the Department of Computer Science, City University of Hong Kong. He has proposed more than 50 technical proposals to ISO/MPEG, ITU-T, and AVS standards, and authored or coauthored more than 200 refereed journal articles/conference papers. He received the Best Paper Award from IEEE VCIP 2019, ICME 2019, IEEE Multimedia 2018, and PCM 2017. His coauthored article received the Best Student Paper Award in the IEEE ICIP 2018. He serves as an Associate Editor for IEEE Transactions on Circuits and Systems for Video Technology. His research interests include video compression, image/video quality assessment, and image/video search and analysis.
\end{IEEEbiography}

\begin{IEEEbiography}[{\includegraphics[width=1in,height=1.25in,clip,keepaspectratio]{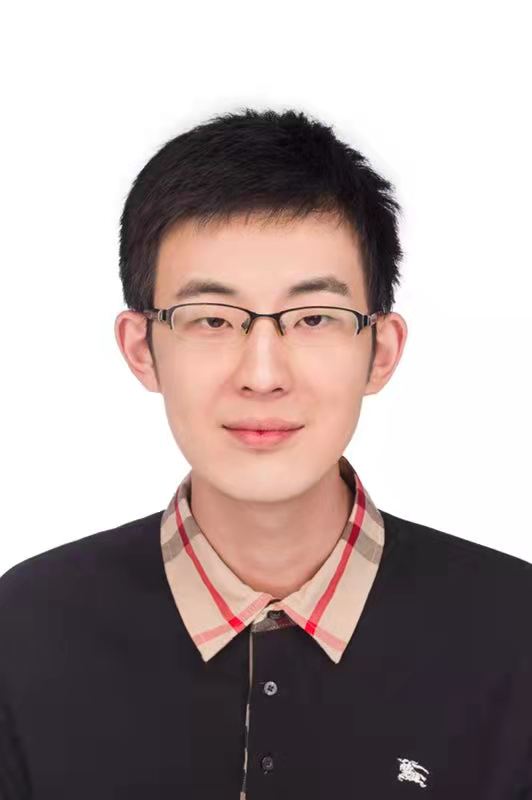}}]
{Lingbo Yang} received the B.S. degree in mathematics in 2016 and the Ph.D. degree in computer application technology in 2021 at Peking University. From 2019 to 2020, he was a research intern at Alibaba DAMO Academy. He is currently working at Tencent Technology (Shanghai) Co. Ltd as a research engineer. His research interests include deep generative models, image restoration and editing, and human pose transfer. 
\end{IEEEbiography}

\begin{IEEEbiography}[{\includegraphics[width=1in,height=1.25in,clip,keepaspectratio]{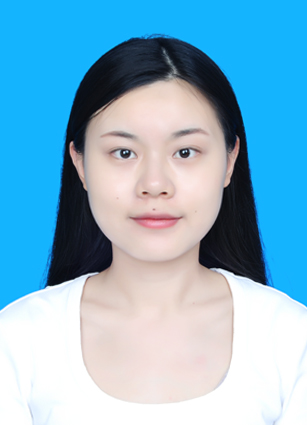}}]
{Qi Mao} received the B.S. degree in digital media technology from Communication University of China in 2016, and PhD degree in EECS at Institute of Digital Media, Peking University in 2021. She is currently an assistant Professor in  School of Information and Communication Engineering and the State Key Laboratory of Media Convergence and Communication, at Communication University of China. 
Her research interests include image/video compression, deep generative models, image synthesis, and content creation.
\end{IEEEbiography}

\begin{IEEEbiography}[{\includegraphics[width=1in,height=1.25in,clip,keepaspectratio]{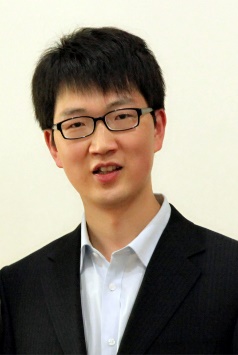}}]
{Jian Zhang} (M'14) received the B.S. degree from the Department of Mathematics, Harbin Institute of Technology (HIT), Harbin, China, in 2007, and received his M.Eng. and Ph.D. degrees from the School of Computer Science and Technology, HIT, in 2009 and 2014, respectively. From 2014 to 2018, he worked as a postdoctoral researcher at Peking University (PKU), Hong Kong University of Science and Technology (HKUST), and King Abdullah University of Science and Technology (KAUST).
Currently, he is an Assistant Professor with the School of Electronic and Computer Engineering, Peking University Shenzhen Graduate School, Shenzhen, China. His research interests include intelligent multimedia processing, deep learning, and optimization. He has published over 80 technical articles in refereed international journals and proceedings. He received the Best Paper Award at the 2011 IEEE Visual Communications and Image Processing (VCIP) and was a co-recipient of the Best Paper Award of 2018 IEEE MultiMedia.
\end{IEEEbiography}

\begin{IEEEbiography}[{\includegraphics[width=1in,height=1.25in,clip,keepaspectratio]{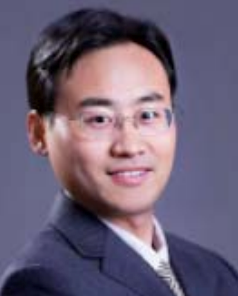}}]
{Siwei Ma} (Senior Member, IEEE) received the B.S. degree from Shandong Normal University, Jinan, China, in 1999, and the Ph.D. degree in computer science from the Institute of Computing Technology, Chinese Academy of Sciences, Beijing, China, in 2005. He held a postdoctoral position with the University of Southern California, Los Angeles, CA, USA, from 2005 to 2007. He joined the School of Electronics Engineering and Computer Science, Institute of Digital Media, Peking University, Beijing, where he is currently a Professor. He has authored over 300 technical articles in refereed journals and proceedings in image and video coding, video processing, video streaming, and transmission. He served/serves as an Associate Editor for the IEEE Transactions on Circuits and Systems for Video Technology and the Journal of Visual Communication and Image Representation.
\end{IEEEbiography}

\end{document}